\documentclass[runningheads]{llncs}
\usepackage{graphicx}
\usepackage{tikz}
\usepackage{comment}
\usepackage{amsmath,amssymb} % define this before the line numbering.
\usepackage{color}

\usepackage{multirow}
\usepackage{subfigure}

\begin{document}
% \linenumbers
\pagestyle{headings}
\mainmatter
\def\ECCVSubNumber{3295}  % Insert your submission number here

\title{Malleable 2.5D Convolution: Learning Receptive Fields along the Depth-axis for RGB-D Scene Parsing} % Replace with your title

\begin{comment}
\titlerunning{ECCV-20 submission ID \ECCVSubNumber}
\authorrunning{ECCV-20 submission ID \ECCVSubNumber}
\author{Anonymous ECCV submission}
\institute{Paper ID \ECCVSubNumber}
\end{comment}
%******************

%\begin{comment}
\titlerunning{Malleable 2.5D Convolution}
\author{Yajie Xing\inst{1}\orcidID{0000-0002-1226-1529} \and
Jingbo Wang\inst{2}\orcidID{0000-0001-9700-6262} \and
Gang Zeng\inst{1}\orcidID{0000-0002-9575-4651}}
\authorrunning{Y. Xing et al.}
\institute{Key Laboratory of Machine Perception, Peking University, China\\
\email{\{yajie\_xing,zeng\}@pku.edu.cn} \and
The Chinese University of Hong Kong \\
\email{jbwang@ie.cuhk.edu.hk}
}
%\end{comment}
%******************
\maketitle

\begin{abstract}
  Depth data provide geometric information that can bring progress in RGB-D scene parsing tasks.
  Several recent works propose RGB-D convolution operators that construct receptive fields along the depth-axis to handle 3D neighborhood relations between pixels.
  However, these methods pre-define depth receptive fields by hyperparameters, making them rely on parameter selection.
  In this paper, we propose a novel operator called malleable 2.5D convolution to learn the receptive field along the depth-axis.
  A malleable 2.5D convolution has one or more 2D convolution kernels.
  Our method assigns each pixel to one of the kernels or none of them according to their relative depth differences, and the assigning process is formulated as a differentiable form so that it can be learnt by gradient descent.
  The proposed operator runs on standard 2D feature maps and can be seamlessly incorporated into pre-trained CNNs.
  We conduct extensive experiments on two challenging RGB-D semantic segmentation dataset NYUDv2 and Cityscapes to validate the effectiveness and the generalization ability of our method.
\keywords{RGB-D Scene Parsing, Geometry in CNN, Malleable 2.5D Convolution}
\end{abstract}

\section{Introduction}
Recent progresses\cite{FCN_PAMI,DeepLabv2,DeepLabv3} in CNN have achieved great success in scene parsing tasks such as semantic segmentation.
Depth data provide geometric information that is not captured by the color channels and therefore can assist feature extraction and improve segmentation performance.
With the availability of commercial RGB-D sensors such as Kinect, there comes an increasing interest in exploiting the additional depth data and incorporating geometric information into CNNs.
% However, standard 2D CNNs' capability to handle geometric information is limited because 2D convolutions only sample pixels according to neighborhood relations on the image.
% Incorporating geometric information into CNNs is an important yet challenging problem and is attracting increasing attention.

Many works\cite{FCN_PAMI,LSTM-CF,FuseNet,ChengCLZH17,RDFNet} take RGB images and depth maps or HHA\cite{HHA} encodings as two separate inputs and adopt two-stream style networks to process them.
These methods only take depth information as features and keep the fixed geometric structures of 2D CNN, which neglects the available 3D geometric relations between pixels.
The most direct method to leverage the 3D relations between pixels is to project RGB-D data into 3D pointclouds\cite{3DGNN} or volumes\cite{SongYZCSF17,ZhongDL18}, and exploit 3D networks to handle geometry.
However, 3D networks are computationally more expensive than 2D CNN and cannot benefit from prevalent ImageNet-pretrained models.
% It is needed to seek a more efficient and flexible method that is able to be incorporated into 2D pretrained CNNs.

Recently, some works turn to introduce geometric information into the 2D convolution operators.
Depth-aware CNN\cite{DepthAware} augments the standard convolution with a depth similarity term.
It applies masks to suppress the contribution of pixels whose depths are different from the center of the kernel.
Consequently, it constructs a soft receptive field along the depth-axis where more distant pixels are partially occluded by the mask.
2.5D convolution\cite{2_5D} moves a step forward to utilizing more kernels to capture richer geometric relations.
It builds a grid receptive field in 3D space, assigns each pixel to one of the kernels according to the relative depth differences with the center of the kernel, and thus mimics a real 3D convolution kernel.
These methods successfully leverage geometric information by introducing a receptive field along the depth-axis.
And they can be easily incorporated into pre-trained CNNs while not bring much computational cost.
However, in these methods, the depth receptive fields are determined by pre-defined hyperparameters.

\begin{figure}[tbp]
  \centering
  \includegraphics[width=.85\textwidth]{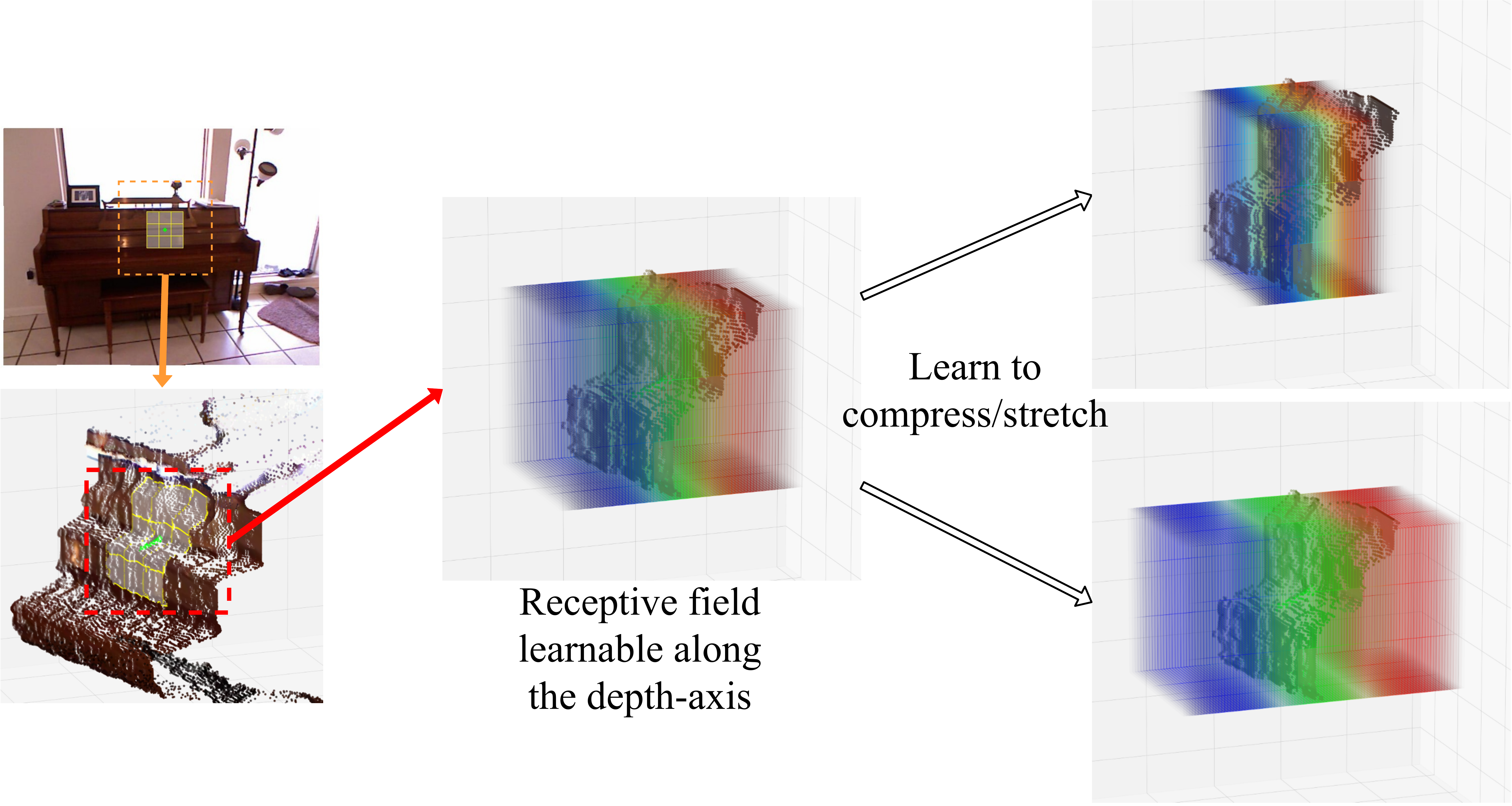}
  \caption{
  Illustration of the malleable 2.5D convolution with 3 kernels.
  Best view in color.
  The blue, green, and red colors respectively represent the receptive field of each kernel.
  The malleable 2.5D convolution arranges its kernels sequentially along the depth-axis and adopts differentiable functions to construct soft depth receptive fields for its kernels.
  It samples pixels on the 2D plane, the same as standard 2D convolution.
  And it assigns pixels to its kernels according to depth data and its depth receptive fields.
  During the training process, the depth receptive fields can be learnt to compress or stretch automatically according to the dataset, making the convolution "malleable"
  }
  \label{fig:idea_illustration}
\end{figure}

In different environments, the scene structures and depth quality can be very different.
For example, NYUDv2\cite{NYUDv2} consists of indoor scenes and captures depth data by Kinect, which has good accuracy.
Cityscapes\cite{Cityscapes} uses a stereo camera to capture outdoor street scenes, resulting in a much longer depth range and noisier depth data.
Naturally, the receptive field along the depth-axis should not be the same across different environment settings.
If we artificially pre-define depth receptive fields for different environments, it would bring many parameter-adjusting works, and the selected parameters are possibly still not suitable for the specific environment.
Therefore, we need a method that can not only build a receptive field along the depth-axis to handle 3D geometry, but also flexibly learn the receptive field for different environments.

To address the aforementioned problems, in this paper, we propose a novel convolution operator called malleable 2.5D convolution (illustrated in Fig.~\ref{fig:idea_illustration}).
Similar to 2.5D convolution, the malleable 2.5D convolution can have either one or more 2D convolution kernels sequentially arranged along the depth-axis.
To determine the depth receptive field of each kernel, we adopt a softmax classification to assign each pixel to kernels according to pixels' relative depth differences.
The assigning process is differentiable and can be learnt by gradient descent.
We also introduce learnable "kernel rebalancing weights" parameters to rebalance the output scale of each kernel, since pixels are not distributed evenly in each class.
The malleable 2.5D convolution can flexibly learn the depth receptive field for different environments (as shown in Fig.~\ref{fig:receptive_field_compare} while only introducing a small number of additional parameters ($2k+3$ parameters if it has $k$ kernels).
Meanwhile, because malleable 2.5D convolutions are based on 2D convolution kernels, they can be easily incorporated into pre-trained 2D CNNs by simply replacing standard 2D convolutions.

Our contributions can be summarized as follows:
\begin{itemize}
  \item We propose a novel convolution operator called malleable 2.5D convolution that has learnable receptive fields along the depth-axis.
  \item Two techniques are proposed in the malleable 2.5D convolution: 1) We propose a differentiable pixel assigning method to achieve learnable depth receptive field;
  2) We introduce "kernel rebalancing weights" parameters to rebalance the uneven pixel distribution in the kernels.
  \item We conduct extensive experiments on both indoor RGB-D semantic segmentation dataset NYUDv2\cite{NYUDv2} and outdoor dataset Cityscapes\cite{Cityscapes}, and validate the effectiveness and generalization ability of our method.
\end{itemize}

\section{Related Works}
\subsubsection{RGB-D Scene Parsing}
Benefiting from the great success of deep convolutional networks\cite{AlexNet,VGGNet,ResNet}, Fully Convolutional Networks (FCNs)\cite{FCN} and its successors\cite{DeepLabv2,RefineNet,DeepLabv3plus,SegModel,PSPNet,DilatedNet} have achieved promising results for RGB semantic segmentation.
RGB-D segmentation extends RGB semantic segmentation by providing additional depth data.
A widely applied method is to encode depth into HHA features\cite{HHA}: horizontal disparity, height above the ground and angle with gravity direction.
It is usually used in two-stream style networks\cite{FCN_PAMI,HHA,LSTM-CF,FuseNet,RDFNet,Coupling} to process RGB images and HHA images and fuse the features or predictions.
Other methods attempt to exploit geometric clues from depth data instead of treating them as features.
Some works\cite{SongYZCSF17,ZhongDL18,3DGNN} transform RGB-D images into 3D data and use 3D networks to handle geometry.
Other works~\cite{CFN,3DN,KongF18,mm2018-KangLN18} take advantage of the fact that the scales of objects in images are inversely proportional to the depths, and change the receptive fields of convolutions according to the depth information.

Depth-aware CNN\cite{DepthAware} and 2.5D Convolution\cite{2_5D} are two methods most related to our work.
Depth-aware CNN applies a mask to construct a soft receptive field along the depth-axis where more distant pixels are weakened and nearer pixels have more contribution to the output.
2.5D convolution seeks to mimic a 3D convolution kernel with several 2D convolution kernels on 2D plane.
It builds a grid receptive field in 3D space, and accordingly assigns each pixel to one of the kernels in a similar way a 3D convolution does.
Both methods pre-define the receptive field along the depth-axis by hyperparameters, making them rely on parameter adjusting and therefore more difficult to be applied in different environments.
Our method solves this problem by building a learnable depth receptive field, and yields better effectiveness and generalization ability.

\subsubsection{Convolutions with Learnable Receptive Fields}
In 2D CNNs, there are several works attempting to break the fixed grid kernel structure of convolution and construct learnable receptive fields.
Spatial Transformer Networks\cite{nips15-STN} warps feature maps with a learned global spatial transformation.
SAC\cite{SAC} learns the scale of receptive fields according to the contents of local features.
PAC\cite{TPAMI2020-ZhangTZLY20} moreover learns convolution kernel shapes according to the perspective.
Deformable convolution\cite{cvpr2017-Deformable,cvpr2019-Deformablev2} learns 2D offsets and adds them to the regular grid sampling locations in the standard convolution.
These works change their receptive field on the 2D plane, which can be implemented by linear transformation and interpolation.
However, along the depth-axis, we are facing a very different problem.
We cannot determine a sample point and then calculate the corresponding feature because pixels in 3D space are very sparse.
We have to do it inversely, assigning pixels to sample points.
And that makes the process more difficult to be differentiable.
Despite the difficulty, our method manages to construct a learnable receptive field along the depth-axis in a novel way.
Another difference is that in this work, we aim to learn the depth receptive field suitable for the given dataset instead of each pixel.

\subsubsection{Neural Architecture Search}
Some neural architecture search works\cite{nips2018-DPC,iclr2019-DARTS,iclr2019-SNAS,MixConv} try to determine the best convolution kernel shapes (or their combination) for the specific dataset.
They use different methods including searching techniques or differentiable optimization.
Our method can also be considered as searching the kernel shapes of malleable 2.5D convolutions, but we only search the receptive fields along the depth-axis and we achieve it through learning by gradient descent.

\section{Malleable 2.5D Convolution}
In this section, we describe and analyze the proposed malleable 2.5D convolution.
We firstly introduce the method to construct learnable receptive fields along the depth-axis.
Then we introduce the kernel rebalancing mechanism in the operator.
Finally, we give some analysis of malleable 2.5D convolution and present how we integrate it into pre-trained CNN.

\subsection{Learning Receptive Fields along the Depth-axis}
\subsubsection{Relative Depth Differences}
First of all, we define the relative depth differences that are used when we construct receptive fields along the depth-axis.
For a convolution kernel, if input map has a downsampling rate $r_{down}$ to the original input image, and the dilation rate is $r_{dilate}$, then the distance between two neighborhood points sampled by the convolution is $\Delta d_p = r_{down} * r_{dilate}$.
% \begin{equation}
%   \Delta d_p = r_{down} * r_{dilate}.
% \end{equation}
Note that $\Delta d_p$ is a distance on the original input image plane.

For a pixel $p_i$ whose coordinate at the input image is $\mathbf{c}_i$, and the depth at $p_i$ is $\mathbf{d}(\mathbf{c}_i)$,
the coordinate of $p_i$ in 3D space is $\mathbf{\hat{c}}_i = (\mathbf{c}_i - (c_x, c_y)) * \mathbf{d}(\mathbf{c}_i) / f$,
% \begin{equation}
%   \mathbf{\hat{c}}_i = (\mathbf{c}_i - (c_x, c_y)) * \mathbf{d}(\mathbf{c}_i) / f,
% \end{equation}
where $c_x$, $c_y$ are the coordinates of the principal point and $f$ is the focal length.
$c_x$, $c_y$, $f$ are all given by camera parameters (usually two focal lengths $f_x$, $f_y$ are given, but in most cases they are very close and we can approximately treat them as the same value).

If we project the local sampling grid of the convolution kernel around $p_i$ to 3D space, then the distance between two neighborhood points would be $\Delta d_s(\mathbf{c}_i) = \Delta d_p * \mathbf{d}(\mathbf{c}_i) / f$.
% \begin{equation}
%   \Delta d_s(\mathbf{c}_i) = \Delta d_p * \mathbf{d}(\mathbf{c}_i) / f.
% \end{equation}
The grid size along the depth-axis should be the same with the other two directions.
Therefore, we define $\Delta d_s(\mathbf{c}_i)$ as the unit of relative depth difference at image coordinate $\mathbf{c}_i$.
And the relative depth difference between pixel at $\mathbf{c}_j$ and $\mathbf{c}_i$ within the local grid centered at $\mathbf{c}_i$ is given by
\begin{equation}
  d(\mathbf{c}_i, \mathbf{c}_j) = (\mathbf{d}(\mathbf{c}_i)-\mathbf{d}(\mathbf{c}_j))/\Delta d_s(\mathbf{c}_i)
\end{equation}

\subsubsection{Depth Receptive Field Functions}
For each pixel $p_i$ whose coordinate at the input image is $\mathbf{c}_i$, a standard 2D convolution performs a weighted sum within the receptive field around $p_i$:
\begin{equation}
  \mathbf{y}(\mathbf{c}_i) = \sum_{\mathbf{c}_p\in\mathcal{R}_p}\mathbf{w}(\mathbf{c}_p)\cdot\mathbf{x}(\mathbf{c}_i+\mathbf{c}_p),
\end{equation}
where $\mathcal{R}_p$ is the local grid that describes the 2D receptive field around $p_i$ in the input $\mathbf{x}$, and $\mathbf{w}$ is the convolution kernel.
Typically, $\mathcal{R}_p$ is a regular grid defined by kernel size and dilation rate.

A malleable 2.5D convolution has $K$ convolution kernels whose receptive fields are sequentially arranged along the depth-axis.
And the pixels are assigned to the kernels according to depth map $\mathbf{d}$:
\begin{equation}
   \mathbf{y}(\mathbf{c}_i) =
   \sum_{k=1}^K
   \sum_{\mathbf{c}_p\in\mathcal{R}_p}
   g_k(\mathbf{d}(\mathbf{c}_i), \mathbf{d}(\mathbf{c}_i+\mathbf{c}_p))
   \cdot
   \mathbf{w}_k(\mathbf{c}_p)
   \cdot
   \mathbf{x}(\mathbf{c}_i+\mathbf{c}_p),
   \label{eq:sum_kernel}
\end{equation}
where $g_k$ is the assigning function for kernel $k$.
$g_k$ defines the depth receptive field of kernel $k$, and satisfies
\begin{equation}
  \sum_{k=1}^Kg_k(\mathbf{d}(\mathbf{c}_i), \mathbf{d}(\mathbf{c}_i+\mathbf{c}_p)) \leq 1, \forall\mathbf{c}_p\in\mathcal{R}_p.
\end{equation}

\begin{figure}[tbp]
  \centering
  \subfigure[$h_k$]{
  \includegraphics[width=0.35\textwidth]{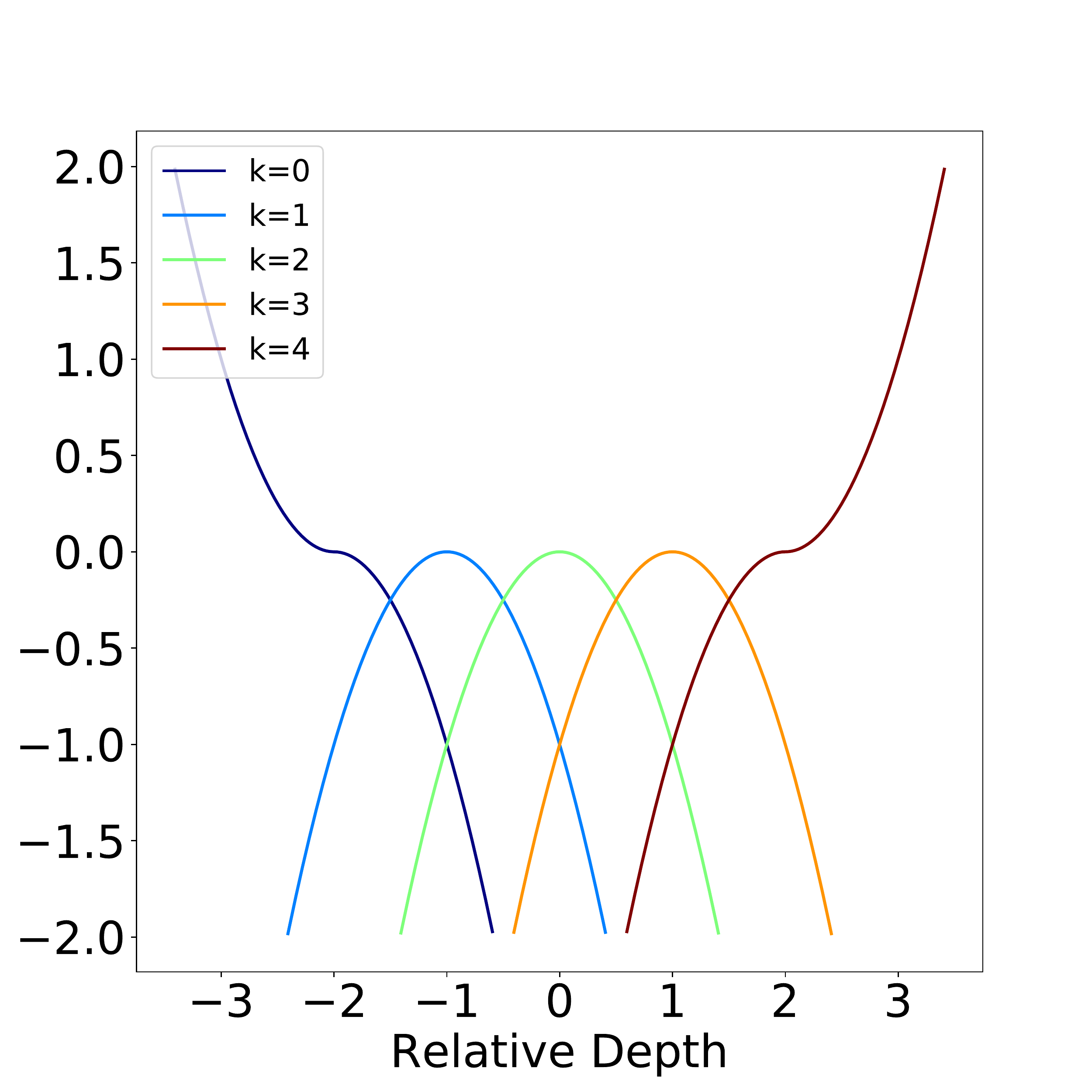}
  }
  \subfigure[$g_k$]{
  \includegraphics[width=0.35\textwidth]{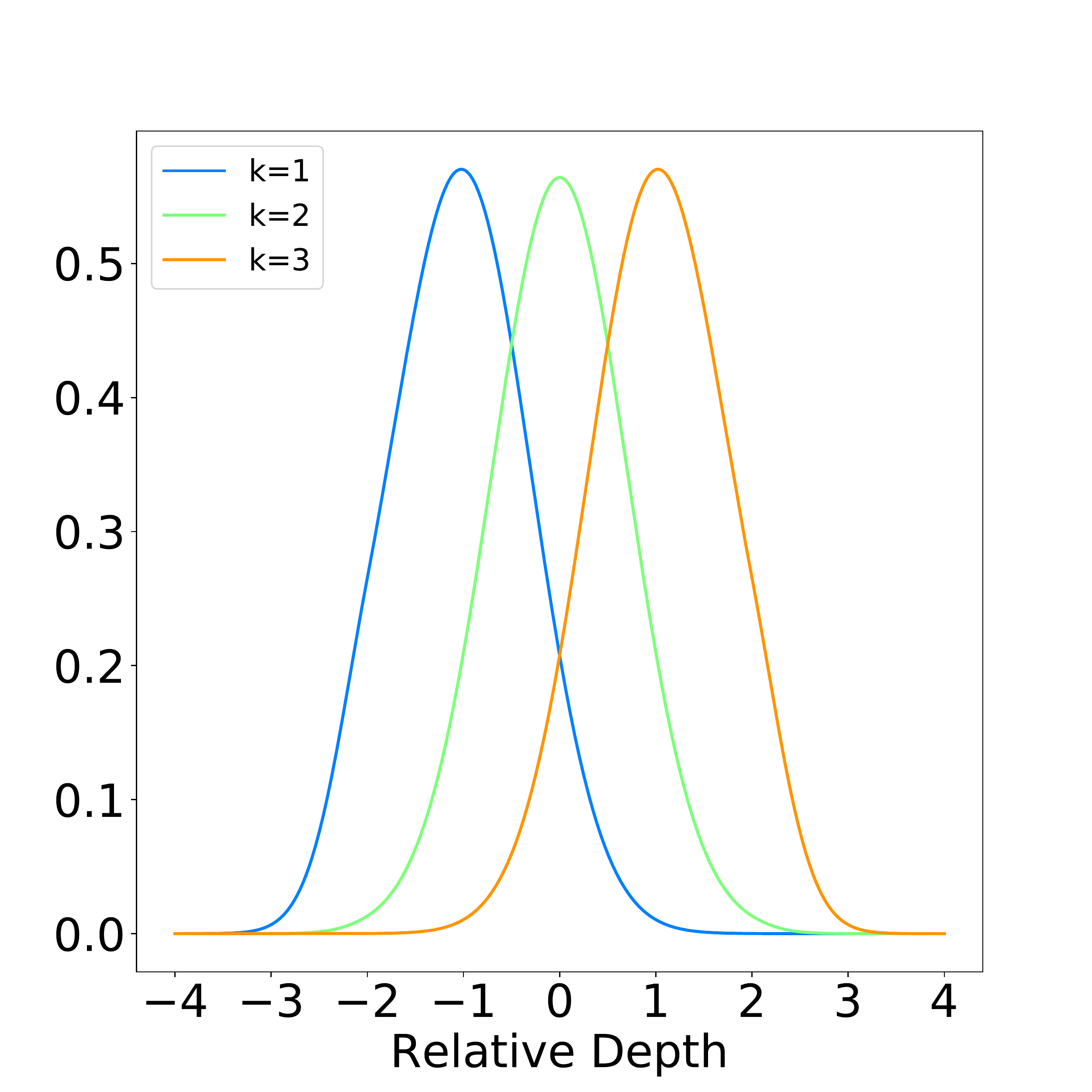}
  }
  \caption{
    Image of $h_k$ and $g_k$ when $K=3$, $[a_0, a_1, a_2, a_3, a_4]=[-2,-1,0,1,2]$ and $t=1$.
    Best view in color
  }
  \label{fig:receptive_field_funcs_eg}
\end{figure}

To make the assigning functions differentiable, we implement them as a softmax classification:
\begin{equation}
  g_k(\mathbf{d}(\mathbf{c}_i), \mathbf{d}(\mathbf{c}_i+\mathbf{c}_p))
  = \frac{exp(h_k(\mathbf{d}(\mathbf{c}_i), \mathbf{d}(\mathbf{c}_i+\mathbf{c}_p)))}
  {\sum_{m=0}^{K+1}exp(h_m(\mathbf{d}(\mathbf{c}_i), \mathbf{d}(\mathbf{c}_i+\mathbf{c}_p))}.
\end{equation}
Here $h_0$ and $h_{K+1}$ are the functions for the two classes outside of all receptive fields (in front of and behind).
For $k=1,2,\cdots,K$, $h_k$ are defined by the relative depth difference as
\begin{equation}
  h_k(\mathbf{d}(\mathbf{c}_i), \mathbf{d}(\mathbf{c}_i+\mathbf{c}_p)) =
  -(d(\mathbf{c}_i, \mathbf{c}_i+\mathbf{c}_p) - a_k)^2/t,
\end{equation}
where $a_k$ is a learnable parameter that determines the center of the kernel's depth receptive field, and $t$ is a learnable temperature parameter that can sharpen/soften the activation of softmax.
$h_0$ and $h_{K+1}$ are defined as
\begin{equation}
  \begin{gathered}
    h_0 = -sgn(d(\mathbf{c}_i, \mathbf{c}_i+\mathbf{c}_p) - a_0)
    \cdot
    (d(\mathbf{c}_i, \mathbf{c}_i+\mathbf{c}_p) - a_0)^2/t, \\
    h_{K+1} = sgn(d(\mathbf{c}_i, \mathbf{c}_i+\mathbf{c}_p) - a_{K+1})
    \cdot
    (d(\mathbf{c}_i, \mathbf{c}_i+\mathbf{c}_p) - a_{K+1})^2/t.
  \end{gathered}
\end{equation}
Here $sgn$ denotes the signum function, which is used to make $h_0$ and $h_{K+1}$ monotonic so that they construct borders of the receptive field.

The constructed depth receptive fields are controlled by parameters $t$ and $a_k, k=0,1,\cdots,K+1$.
And both sets of parameters are learnable by gradient descent.
Fig~\ref{fig:receptive_field_funcs_eg} gives an example of the image of $h_k$ and $g_k$.
% Some images of $h_k$ and $g_k$ will be shown in the supplementary material.

\subsection{Kernel Rebalancing}

\begin{figure}[htbp]
  \centering
  \subfigure[Before rebalance]{
  \includegraphics[width=0.35\textwidth]{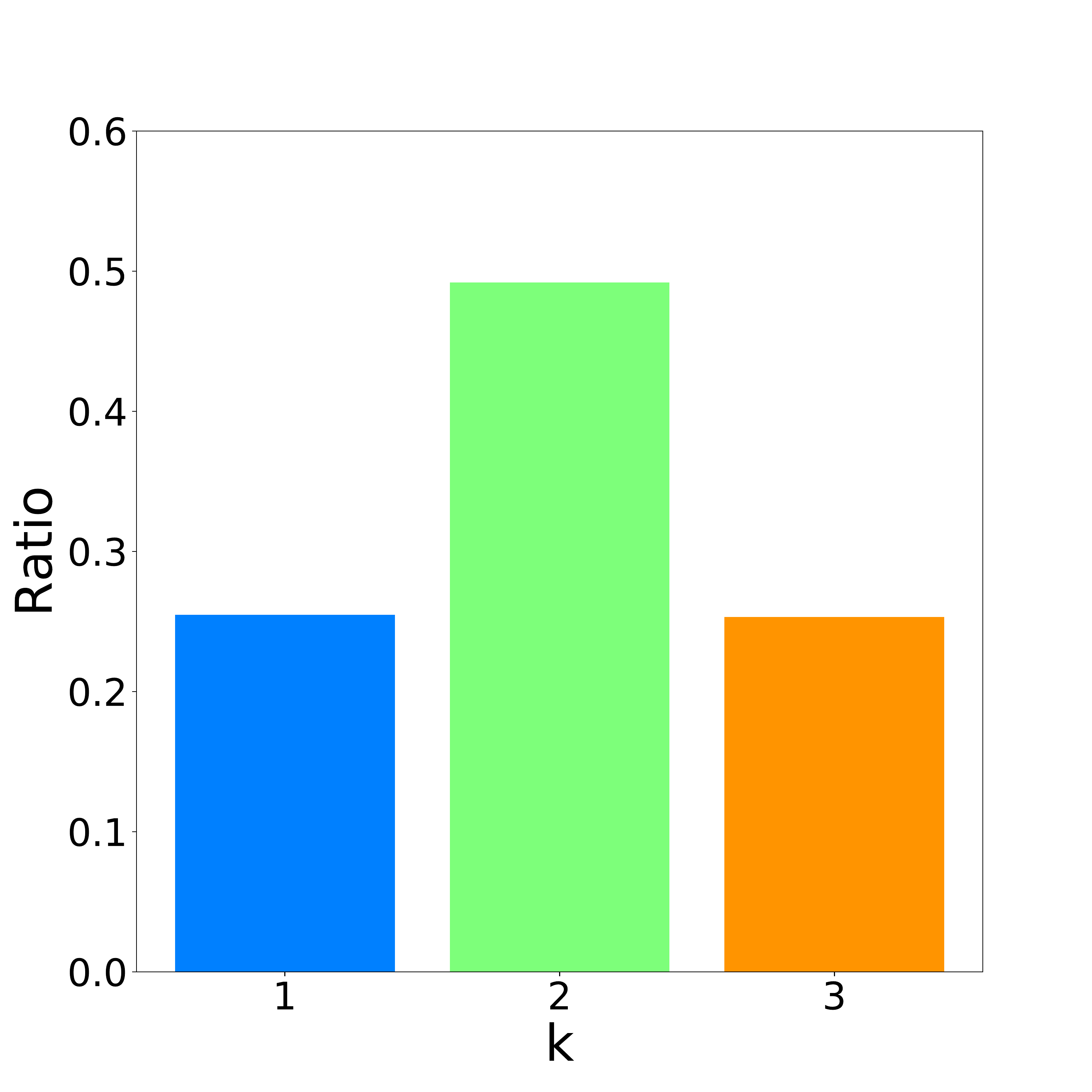}
  }
  \subfigure[After rebalance]{
  \includegraphics[width=0.35\textwidth]{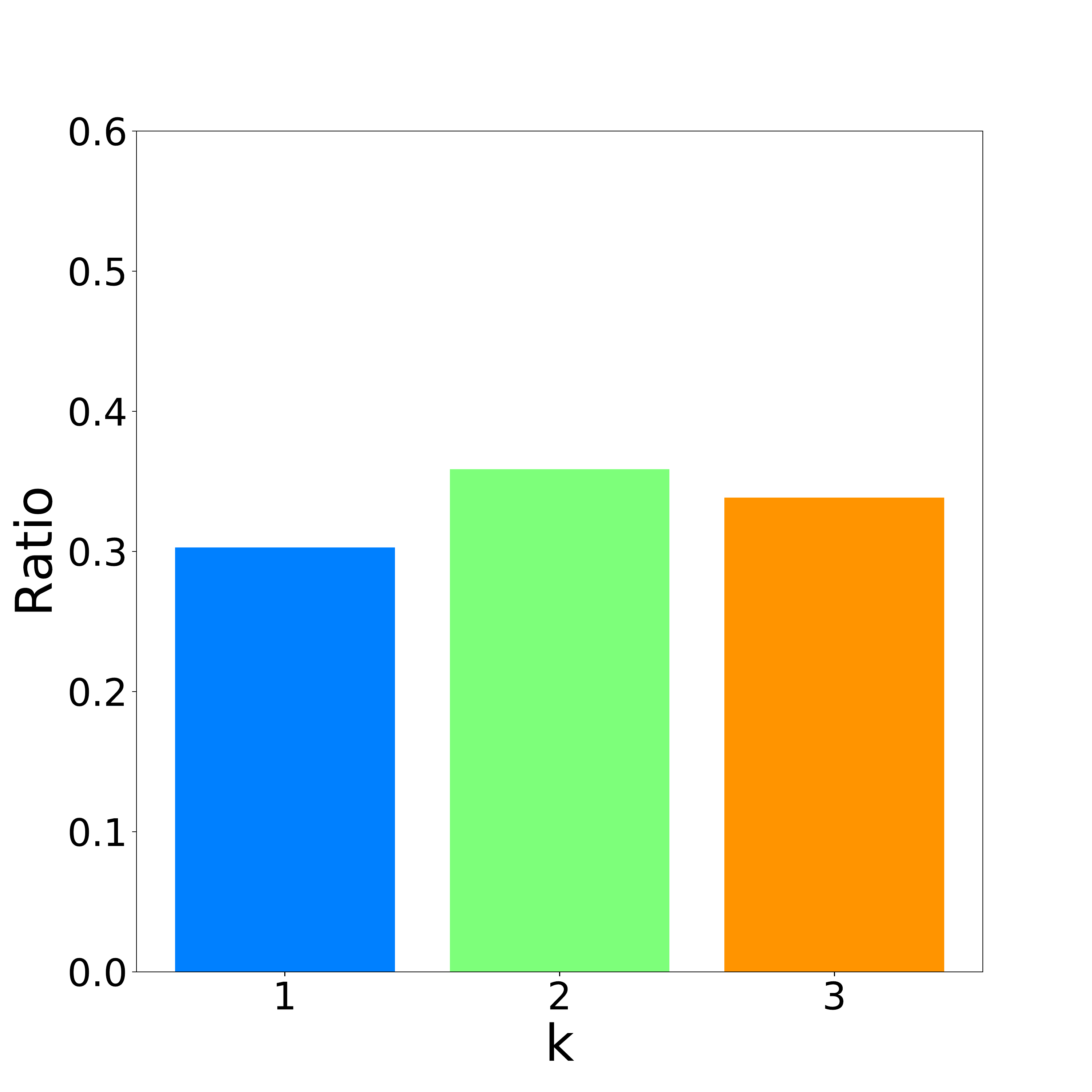}
  }
  \caption{
    The ratio of pixels assigned to each kernel, before and after rebalance.
    The figure shows the malleable 2.5D convolution at res5 stage of a trained model.
    We count the sum of $g_k$ and $s_k\cdot g_k$ for each kernel across the whole NYUDv2 dataset and calculate the ratio.
    More cases will be presented in the supplementary material.
  }
  \label{fig:rebalance}
\end{figure}

Through the receptive field functions $g_k$, we can assign pixels to convolution kernels.
However, pixels are not evenly distributed along the depth-axis.
In other words, the expectation
 $\mathbb{E}[g_k(\mathbf{d}(\mathbf{c}_i), \mathbf{d}(\mathbf{c}_i+\mathbf{c}_p))]$
are not equal for different parameters of $g_k$.
This implies that the outputs of different kernels might have different value scales, and the scales change with parameters of $g_k$.
Theoretically, the scale change can be adjusted by $\mathbf{w}_k$, but implicitly learning a scale factor in convolution kernel weights could increase learning difficulty.
Therefore, we introduce a scale factor $s_k$ to rebalance the output scales of different kernel.
Then Eq.~\ref{eq:sum_kernel} is modified as
\begin{equation}
   \mathbf{y}(\mathbf{c}_i) =
   \sum_{k=1}^K
   \sum_{\mathbf{c}_p\in\mathcal{R}_p}
   s_k \cdot
   g_k(\mathbf{d}(\mathbf{c}_i), \mathbf{d}(\mathbf{c}_i+\mathbf{c}_p))
   \cdot
   \mathbf{w}_k(\mathbf{c}_p)
   \cdot
   \mathbf{x}(\mathbf{c}_i+\mathbf{c}_p),
\end{equation}
And $s_k$-s must satisfy
$s_k \geq 0, \forall k=1,2,\cdots,K$
and
$\sum_{k=1}^K s_k = 1$.
% \begin{equation}
%   \begin{gathered}
%     s_k \geq 0, \forall k=1,2,\cdots,K, \text{  }
%     \sum_{k=1}^K s_k = 1.
%   \end{gathered}
% \end{equation}
Note that $\sum_{k=1}^K s_k$ can actually equal to any constant considering that in modern CNNs, a convolution layer is almost always followed by a normalization layer.
To ensure that $s_k$-s satisfy the conditions, we implement them as
\begin{equation}
  s_k = \frac{exp(b_k)}{\sum_{k=1}^K exp(b_k)}
\end{equation}
And the malleable 2.5D convolution learns $b_k$ by gradient descent.

Fig.~\ref{fig:rebalance} shows the effect of kernel rebalancing.
We can see that the kernel rebalancing mechanism meets our expectation and successfully rebalance the pixel distribution in different kernels.

\subsection{Understanding Malleable 2.5D Convolution}
\subsubsection{Comparisons with other RGB-D convolutions}
To get a better understanding of learnable depth receptive fields, we compare our malleable 2.5D convolution with the two previous RGB-D convolutions, Depth-aware Convolution and the 2.5D Convolution.
% receptive fields
These two convolutions can both be written as the form of Eq.~\ref{eq:sum_kernel}.
In Depth-aware Convolution, $K$ is $1$, and $g_1$ is defined as
\begin{equation}
  g_1(\mathbf{d}(\mathbf{c}_i), \mathbf{d}(\mathbf{c}_i+\mathbf{c}_p))
  =
  exp(-\alpha|\mathbf{d}(\mathbf{c}_i)-\mathbf{d}(\mathbf{c}_i+\mathbf{c}_p)|),
\end{equation}
where $\alpha$ is a pre-defined constant.
And in 2.5D convolution, $K$ can be any positive integer, and $g_k$ is defined as
\begin{equation}
  g_k(\mathbf{d}(\mathbf{c}_i), \mathbf{d}(\mathbf{c}_i+\mathbf{c}_p))
  =
  \left\{
  \begin{aligned}
      &1, & & k-1-\frac{K}{2} \leq d(\mathbf{c}_i, \mathbf{c}_i+\mathbf{c}_p) < k-\frac{K}{2}\\
      &0, & & otherwise
    \end{aligned}
  \right.
\end{equation}

\begin{figure}[htbp]
  \centering
  \includegraphics[width=0.85\textwidth]{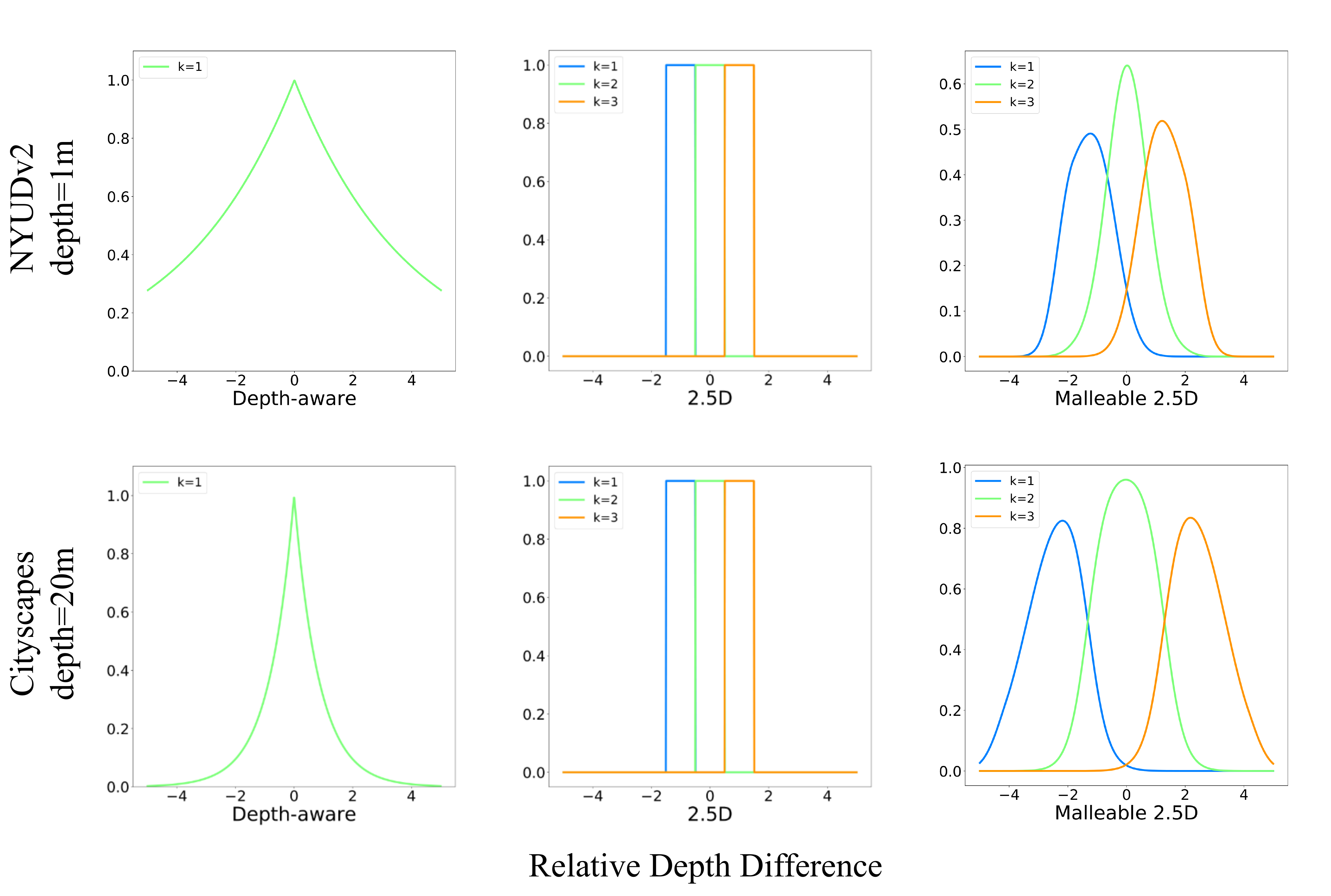}
  \caption{
  Comparison of depth receptive field functions $g_k$.
  Best view in color.
  We compare the depth receptive fields at res5 stage after training.
  The first row shows a case in NYUDv2 where the depth $\mathbf{d}(\mathbf{c}_i)=1m$.
  The second row shows a case in Cityscapes where the depth $\mathbf{d}(\mathbf{c}_i)=20m$.
  Note that we scale the y-axis to see better.
  The overall scale does not affect output results because of batch normalizations
  }
  \label{fig:receptive_field_compare}
\end{figure}

In Fig~\ref{fig:receptive_field_compare}, we compare depth receptive fields of these two convolutions and our method.
We visualize depth receptive fields in two typical cases in NYUDv2 and Cityscapes respectively.
In the two datasets, both focal lengths and the depth ranges are different.
The Depth-aware Convolution uses absolute depth difference to control the receptive field.
When the depths and focal length changes, the 3D distance between two adjacent pixels on the image could be very different.
And we can see that Depth-aware Convolution cannot fit both cases with the same parameter setting.
2.5D convolution adopts relative depth differences, which makes it more robust to varying depth and focal length.
However, 2.5D convolution's receptive field along the depth-axis is also pre-determined and cannot adapt to different datasets without handcraft adjusting.
In contrast, malleable 2.5D convolution not only can avoid the influence of varying depth and focal length, but also automatically learn the depth receptive field for different environments.
As an outdoor dataset, Cityscapes has a larger range of depth.
And the depth maps in Cityscapes are relatively noisy, not as sharp as those in NYUDv2.
Intuitively, the depth receptive field for Cityscapes should be larger and less sharp.
We can see from the figure that malleable 2.5D convolution learns a wider depth receptive field for Cityscapes than that for NYUDv2, and the edges drop slower.
% The experimental results that compares these two operator with our method are presented in Table~\ref{table:nyu_conv_compare} and \ref{table:city_conv_compare}.

\begin{table}[htbp]
  \begin{center}
  \caption{
  Comparison of the computational cost of different convolutions.
  % The baseline is ResNet-101 based DeepLabv3+\cite{DeepLabv3plus}.
  The input size is $768\times 768$.
  % To evaluate different convolutions, we replace the $3\times3$ convolution with a RGB-D convolution in the first residual unit in each stage of the ResNet.
  "kernels" means the kernel number of used RGB-D convolutions
  }
  \label{table:computation_cost}
  \setlength{\tabcolsep}{10pt}
  \begin{tabular}{lccc}
    \hline\noalign{\smallskip}
    Method & kernels & FLOPs(G) & Params(M)\\
    \noalign{\smallskip}
    \hline
    \noalign{\smallskip}
    Baseline                  & 1 & 215.673 & 59.468 \\
    \noalign{\smallskip}
    \hline
    \noalign{\smallskip}
    Depth-aware\cite{DepthAware}    & 1 & 215.675 & 59.468 \\
    Malleable 2.5D                    & 1 & 215.679 & 59.468 \\
    \noalign{\smallskip}
    \hline
    \noalign{\smallskip}
    2.5D\cite{2_5D}           & 3 & 234.701 & 65.734 \\
    Malleable 2.5D              & 3 & 234.711 & 65.734 \\
    \hline
  \end{tabular}
  \end{center}
\end{table}
% computational cost
In Table~\ref{table:computation_cost}, we compare the FLOPs and parameter number of different convolutions.
We present ResNet-based DeepLabv3+\cite{DeepLabv3plus} as the baseline, and replace the $3\times3$ convolution with a RGB-D convolution in the first residual unit in each stage of the ResNet.
It shows that when using the same number of kernels, the malleable 2.5D convolution only brings very minor additional computational cost to achieve learnable depth receptive fields.

\subsubsection{Usage of Malleable 2.5D Convolution}
Malleable 2.5D convolutions can be easily incorporated into CNNs by replacing standard 2D convolutions.
The inputs of malleable 2.5D convolutions are standard feature maps, depth maps, and camera parameters.
And the outputs of malleable 2.5D convolutions are the same with standard 2D convolutions.

In RGB-D semantic segmentation, utilizing pre-trained models is essential to attain good performance.
When replacing standard convolutions with malleable 2.5D convolutions, we do not abandon the pre-trained weights of the original convolutions but adopt a simple parameter loading strategy to make use of them.
We duplicate the pre-trained weights and load them into each kernel of the malleable 2.5D convolution.
in finetuning time, the $k$ kernels start from the same initialization and gradually learn to model geometric relations.

\section{Experiments}
\subsection{Datasets}
We evaluate our method on two popular RGB-D scene parsing datasets: NYUDv2\cite{NYUDv2} and Cityscapes\cite{Cityscapes}.
These two datasets respectively contains indoor and outdoor scenes and the depth sources are different.
Therefore, the depth ranges, object sizes and the quality of depth data are very different in them.
Evaluating on these two datasets is challenging and can validate the robustness and generalization ability of RGB-D scene parsing methods.
We will introduce these two datasets in detail in the supplementary material.

% \subsubsection{NYUDv2}
% NYUDv2\cite{NYUDv2} is an indoor RGB-D semantic segmentation dataset.
% It contains 1449 RGB-D images with pixel-wise labels.
% And it provides depth maps captured by Kinect and the corresponding camera intrinsic parameters for all images.
% We follow the 40-class setting and the standard split which consists of 795 training images and 654 testing images.
% \subsubsection{Cityscapes}
% Cityscapes\cite{Cityscapes} is an urban scene understanding dataset that contains outdoor scene in different cities.
% The dataset has 5,000 stereo frames,
% each frame containing an $2048\times1024$ RGB image, a disparity map, a set of camera parameters, and a fine-annotated 19-category ground truth label map.
% There are 2,979 images in training set, 500 images in validation set and 1,525 images in test set.
% We use camera parameters and disparity maps to calculate depth maps.
% The quality of depth data in this dataset is not as good as NYUDv2, and the scenes have wider ranges and more complicated structures.

\subsection{Implementation Details}
\subsubsection{Model Implementation}
DeepLabv3+\cite{DeepLabv3plus} is a widely recognized state-of-the-art method for semantic segmentation.
% And ResNet\cite{ResNet} is the most commonly used backbone model.
We adopt ResNet-based DeepLabv3+ pre-trained on ImageNet\cite{ImageNet} as our baseline network.
And for NYUDv2, we moreover adopt a multi-stage merging block inspired by previous works\cite{RefineNet,RDFNet}.
% For the NYUDv2 dataset, inspired by previous works\cite{miccai2015-UNet,RefineNet,RDFNet} that merge multi-stage features to enhance detailed prediction, we moreover adopt a multi-stage merging block on the backbone network.
% And for Cityscapes, we keep the original DeepLabv3+ structure.
Details of the network structures will be illustrated in the supplementary material.
To evaluate the our method, we replace the $3\times3$ convolution with a malleable 2.5D convolution in the first residual unit in each stage of the ResNet.

Our implementation is based on PyTorch\cite{PyTorch}.
Synchronized batch normalization are adopted for better batch statistics.
By default, we use malleable 2.5D convolution with 3 kernels, and the parameter $[a_0, a_1, a_2, a_3, a_4]$ are initialized as $[-2,-1,0,1,2]$, $t$ is initialized as $1$, and $b_k$-s are all initialized as $0$.

\subsubsection{Training Settings}
We use SGD optimizer with momentum to train our model. The momentum is fixed as 0.9 and the weight decay is set to 0.0001. We employ a "poly" learning rate policy where the initial learning rate is multiplied by $(1-\frac{iter}{max\_iter})^{power}$. The initial learning rate is set to 0.01 and the power is set to 0.9.
We use batch size of 16 and train our model for 40k iterations for NYUDv2 and 60k iterations for Cityscapes.
For data augmentation, we use random cropping, random horizontal flipping and random scaling with scale $\in\{0.75,1,1.25,1.5,1.75,2\}$.
% And we randomly crop the input image to $480\times480$ for NYUDv2 and $800\times800$ for Cityscapes.
Besides, we adopt the bootstrapped cross-entropy loss as in \cite{arxiv2016-WuSH16a} for Cityscapes experiments.

\subsection{Main Results}
\begin{table}[tbp]
  \begin{center}
  \caption{Comparison with other RGB-D convolutions on NYUDv2 and Cityscapes.
  The backbone model is ResNet-50.
  "kernels" means the kernel number of used RGB-D convolutions
  }
  \label{table:conv_compare}
  \setlength{\tabcolsep}{5.5pt}
  \begin{tabular}{lccccc}
    \hline\noalign{\smallskip}
    \multirow{2}{*}{Method} & \multirow{2}{*}{kernels} &
    \multicolumn{2}{c}{NYUDv2} & \multicolumn{2}{c}{Cityscapes} \\
     & & mIoU(\%) & pixel Acc(\%) & mIoU(\%) & pixel Acc(\%)\\
    \noalign{\smallskip}
    \hline
    \noalign{\smallskip}
    Baseline      & 1 & 44.56  & 73.01 & 79.94  & 96.34\\
    \noalign{\smallskip}
    \hline
    \noalign{\smallskip}
    Depth-aware\cite{DepthAware}   & 1 & 46.69 & 74.27 & 79.01 & 96.32 \\
    Malleable 2.5D  & 1 & \textbf{47.08} & \textbf{75.13} & \textbf{80.26} & \textbf{96.40} \\
    \noalign{\smallskip}
    \hline
    \noalign{\smallskip}
    2.5D\cite{2_5D}          & 3 & 48.23 & 75.73 & 78.63 & 96.29 \\
    Malleable 2.5D  & 3 & \textbf{48.80} & \textbf{76.03} & \textbf{80.81} & \textbf{96.51}\\
    \hline
  \end{tabular}
  \end{center}
\end{table}

\begin{table}[htbp]
  \begin{center}
  \caption{Comparison with state-of-the-art RGB-D scene parsing methods on NYUDv2.
  Multi-scale and flipping inference strategies are used when evaluating our method.
  % Note that some state-of-the-art methods use RGB+HHA inputs and adopt the two-stream network design that doubles the backbone
  }
  \label{table:nyu_sota_compare}
  \setlength{\tabcolsep}{9pt}
  \begin{tabular}{lccc}
    \hline\noalign{\smallskip}
    Method & Backbone & mIoU(\%) & pixel Acc(\%)\\
    \noalign{\smallskip}
    \hline
    \noalign{\smallskip}
    FCN+HHA\cite{FCN_PAMI} & VGG-16 $\times2$ & 34.0 & 65.4 \\
    % 3DGNN+HHA\cite{3DGNN} & ResNet-101 $\times2$ & 43.1 & - \\
    Depth-aware+HHA\cite{DepthAware} & VGG-16 $\times2$ & 43.9 & - \\
    % Kong \textit{et al.}\cite{KongF18} & ResNet-101 $\times1$ & 44.5 & 72.1 \\
    CFN(RefineNet-152)\cite{CFN} & ResNet-152 $\times2$ & 47.7 & -\\
    % 3DN\cite{3DN} & ResNet-101 $\times1$ & 48.2 & 74.8\\
    2.5D\cite{2_5D} & ResNet-101 $\times1$ & 48.4  & 75.3 \\
    2.5D+HHA\cite{2_5D} & ResNet-101 $\times2$ & 49.1  & 75.9 \\
    RDF-101\cite{RDFNet} & ResNet-101 $\times2$ & 49.1 & 75.6\\
    RDF-152\cite{RDFNet} & ResNet-152 $\times2$ & 50.1 & 76.0 \\
    Idempotent\cite{Coupling} & ResNet-101 $\times2$ & 50.6 & 76.3 \\
    \noalign{\smallskip}
    \hline
    \noalign{\smallskip}
    Malleable 2.5D   & ResNet-50 $\times1$ & 49.7 & 76.3 \\
    Malleable 2.5D  & ResNet-101 $\times1$ & \textbf{50.9} & \textbf{76.9} \\
    \hline
  \end{tabular}
  \end{center}
\end{table}

\begin{figure}[tbp]
  \centering
  \includegraphics[width=.82\textwidth]{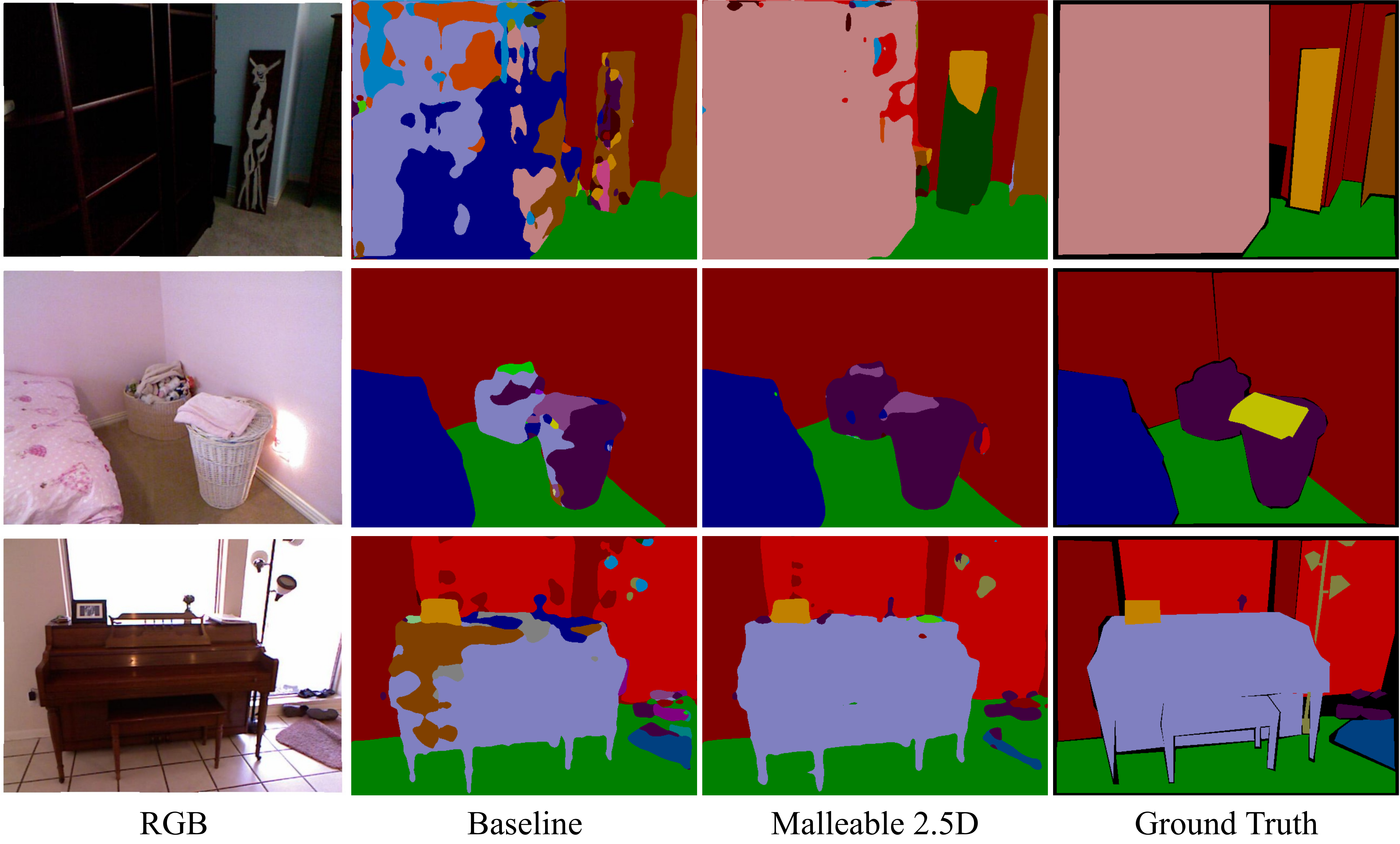}
  \caption{
  Segmentation results on NYUDv2 test dataset.
  Black regions in "Ground Truth" is are the ignoring category.
  Models are based on ResNet-50
  }
  \label{fig:results_vis}
\end{figure}

First of all, we compare our method with previous RGB-D convolutions to validate the benefits of learnable receptive fields along the depth-axis.
The results are shown in Table~\ref{table:conv_compare}.
We compare Depth-aware, 2.5D and malleable 2.5D convolution on both NYUDv2 and Cityscapes datasets.
When implementing the other two methods, we use the default parameters provided in their papers\cite{DepthAware,2_5D} which are tuned on NYUDv2.
All evaluations are conducted without multi-scale and flip testing tricks.
For fairness, we compare malleable 2.5D convolution with other convolutions in the case that they have the same kernel number.
And the single-kernel version of malleable 2.5D convolution's parameter $[a_0, a_1, a_2]$ are initialized as $[-1,0,1]$.
The results show that the malleable 2.5D convolution consistently outperforms other methods and baselines.
It is worth noting that Depth-aware and 2.5D convolutions fail on Cityscapes.
They decrease the performance even depth information is incorporated.
However, our malleable 2.5D convolution still works well on Cityscapes with the same parameter initialization on NYUDv2.
It adaptively learns the depth receptive field that fits the outdoor environment and improves segmentation performance.
Fig.~\ref{fig:results_vis} visualizes qualitative comparison results between the baseline and our method on NYUv2 test set.

NYUDv2 provides high-quality depth data, and most RGB-D scene parsing methods take it as the main benchmark.
Therefore, we then compare with more state-of-the-art RGB-D scene parsing methods on NYUDv2, shown in Table~\ref{table:nyu_sota_compare}.
As other works do\cite{RDFNet,2_5D,Coupling}, we adopt test-time multi-scale and flip inference to obtain best performance.
We evaluate models based on both ResNet-50 and ResNet-101 backbones.
And we use malleable 2.5D convolutions with 3 kernels.
Different from many works that adopt two-stream network design to process RGB+HHA inputs, we do not use HHA input and thus do not double the computational cost brought by backbones.
The results show that even without doubling the backbone, our model achieves promising performance and outperforms other methods.

\subsection{Ablation Studies}
In this subsection, we conduct ablation studies on NYUDv2 dataset to validate the efficacy of our method.

\subsubsection{Number of kernels}
In Table~\ref{table:ablation_kernel_nums}, we conduct experiments that compare different numbers of convolutional kernels in malleable 2.5D convolutions.
While the performance improves when the number increases from 1 to 3, the performance of the 5-kernel case drops slightly compared to the 3-kernel case.
We suppose the reason is that 3 kernels are enough to model 3D relations between pixels, and in the 5-kernel case there are too few pixels assigned to the two outer kernels and they cannot be sufficiently trained.

\begin{table}[htbp]
  \begin{center}
  \caption{
  Ablation study for number of convolution kernels.
  The backbone is ResNet-50
  }
  \label{table:ablation_kernel_nums}
  \setlength{\tabcolsep}{7pt}
  \begin{tabular}{ccc}
    \hline\noalign{\smallskip}
    Kernel numbers & mIoU(\%) & pixel Acc(\%)\\
    \noalign{\smallskip}
    \hline
    \noalign{\smallskip}
    1 & 47.08 & 75.13 \\
    3 & \textbf{48.80} & \textbf{76.03} \\
    5 & 48.17 & 75.59 \\
    \hline
  \end{tabular}
  \end{center}
\end{table}

\subsubsection{Learnable parameters}
In Table~\ref{table:ablation_learnable_params}, we evaluate the effects of introduced learnable parameters.
We fix part of the introduced parameters \{$a_k$, $t$, $b_k$\} at the initialized values, and set rest of them learnable.
The results demonstrate that when we fix all the parameters and thus fix the depth receptive fields, the performance is the worst.
And the fixed malleable 2.5D convolution gets a very similar result to 2.5D convolution.
When only the kernel center $a_k$ is learnable, the performance stays almost the same.
And when both $a_k$ and $t$ are learnable, the mIoU improves from 48.33\% to 48.62\% and pixel accuracy improves from 75.74\% to 75.91\%.
We argue that this is because the temperature term T controls the sharpness of depth receptive fields, and therefore it is crucial to set both $a_k$ and $t$ learnable for assigning pixels to different kernels more accurately.
When the kernel rebalancing parameter $b_k$ is introduced, the performance moreover improves to the result of complete malleable 2.5D convolution.
These results validate that the improvement is indeed brought by the introduced learnable parameters.

\begin{table}[htbp]
  \begin{center}
  \caption{
  Ablation study for introduced learnable parameters in malleable 2.5D convolution.
  The backbone model is ResNet-50
  }
  \label{table:ablation_learnable_params}
  \setlength{\tabcolsep}{10pt}
  \begin{tabular}{llcc}
    \hline\noalign{\smallskip}
    Method & Learnable params & mIoU(\%) & pixel Acc(\%)\\
    \noalign{\smallskip}
    \hline
    \noalign{\smallskip}
    Baseline      & - & 44.56  & 73.01 \\
    2.5D\cite{2_5D}          & - & 48.23 & 75.73 \\
    \noalign{\smallskip}
    \hline
    \noalign{\smallskip}
    \multirow{4}{*}{Malleable 2.5D} & None              & 48.28 & 75.68 \\
     & $a_k$             & 48.33 & 75.74 \\
     & $a_k$, $t$        & 48.62 & 75.91 \\
     & $a_k$, $t$, $b_k$ & \textbf{48.80} & \textbf{76.03} \\
    \hline
  \end{tabular}
  \end{center}
\end{table}

\subsubsection{Initialization of introduced parameters}
The parameter $a_k$ serves as the center of kernel $k$'s receptive field along the depth-axis.
It is the most direct factor that controls the learnt depth receptive field.
In Table~\ref{table:ablation_initialization}, we evaluate the effects of different initialization of $a_k$ for 3-kernel malleable 2.5D convolution.
When we use large values to initialize $a_k$, the performance drops slightly.
However, both results outperform the baseline and 2.5D convolution.
This validates the effectiveness and the robustness of the learnable depth receptive field in our method.
More experiments of different parameter initializations will be included in the supplementary material.

\begin{table}[htbp]
  \begin{center}
  \caption{
  Results of different initialization of $a_k$.
  The backbone model is ResNet-50
  }
  \label{table:ablation_initialization}
  \setlength{\tabcolsep}{10pt}
  \begin{tabular}{llccc}
    \hline\noalign{\smallskip}
    Method & Initialization & mIoU(\%) & pixel Acc(\%)\\
    \noalign{\smallskip}
    \hline
    \noalign{\smallskip}
    Baseline      & - & 44.56  & 73.01 \\
    2.5D\cite{2_5D}          & - & 48.23 & 75.73 \\
    \noalign{\smallskip}
    \hline
    \noalign{\smallskip}
    \multirow{2}{*}{Malleable 2.5D} & $[-4,-2,0,2,4]$ & 48.66 & 75.94 \\
     & $[-2,-1,0,1,2]$ & \textbf{48.80} & \textbf{76.03} \\
    \hline
  \end{tabular}
  \end{center}
\end{table}

\subsubsection{Malleable 2.5D Convolution in Different Layers}
To reveal the effects of using malleable 2.5D convolution in different locations in the backbone network, we conduct a series of experiments.
The results are shown in Table~\ref{table:ablation_location}.
When using more malleable 2.5D convolution, the network gains more capability to handle 3D geometric information and achieves better segmentation performance.

\begin{table}[htbp]
  \begin{center}
  \caption{
  Results of using malleable 2.5D convolutions in different layers.
  "Replaced Location" means in which stage in the ResNet the first $3\times 3$ convolution is replaced by malleable 2.5D convolution.
  The backbone model is ResNet-50
  }
  \label{table:ablation_location}
  \setlength{\tabcolsep}{10pt}
  \begin{tabular}{llccc}
    \hline\noalign{\smallskip}
    Method & Replaced Location & mIoU(\%) & pixel Acc(\%)\\
    \noalign{\smallskip}
    \hline
    \noalign{\smallskip}
    Baseline      & - & 44.56  & 73.01 \\
    \noalign{\smallskip}
    \hline
    \noalign{\smallskip}
    \multirow{4}{*}{Malleable 2.5D} & res2, res3 & 48.27 & 75.83 \\
     & res4, res5             & 48.34 & 75.77 \\
     & res3, res4, res5       & 48.58 & 75.93 \\
     & res2, res3, res4, res5 & \textbf{48.80} & \textbf{76.03} \\
    \hline
  \end{tabular}
  \end{center}
\end{table}

\section{Conclusion}
We propose a novel RGB-D convolution operator called malleable 2.5D convolution, which has learnable receptive fields along the depth-axis.
% A malleable 2.5D convolution consists of one or several 2D convolution kernels that are sequentially arranged along the depth-axis to model 3D geometric relations between pixels.
% It includes differentiable pixel assigning functions that yield learnable depth receptive fields, and a set of kernel rebalancing parameters to handle the uneven pixel distribution in the kernels.
By learning the receptive field along the depth-axis, malleable 2.5D convolution can learn to adapt to different environments without handcraft parameter adjusting and improve RGB-D scene parsing performance.
And malleable 2.5D convolution can be easily incorporated into pre-trained standard CNNs.
Our extensive experiments on NYUDv2 and Cityscapes validate the effectiveness and the generalization ability of our method.
\\
\\
\noindent \textbf{Acknowledgments:} This work is supported by the National Key Research and Development Program of China (2017YFB1002601, 2016QY02D0304), National Natural Science Foundation of China (61375022, 61403005, 61632003), Beijing Advanced Innovation Center for Intelligent Robots and Systems (2018IRS11), and PEK-SenseTime Joint Laboratory of Machine Vision.

\bibliographystyle{splncs04}
\bibliography{ref_pool}

\clearpage
\section*{Appendix}
\setcounter{section}{0}

\section{Datasets}
\subsubsection{NYUDv2}
NYUDv2\cite{NYUDv2} is an indoor RGB-D semantic segmentation dataset.
It contains 1449 RGB-D images with pixel-wise labels.
And it provides depth maps captured by Kinect and the corresponding camera intrinsic parameters for all images.
We follow the 40-class setting and the standard split which consists of 795 training images and 654 testing images.
\subsubsection{Cityscapes}
Cityscapes\cite{Cityscapes} is an urban scene understanding dataset that contains outdoor scene in different cities.
The dataset has 5,000 stereo frames,
each frame containing an $2048\times1024$ RGB image, a disparity map, a set of camera parameters, and a fine-annotated 19-category ground truth label map.
There are 2,979 images in training set, 500 images in validation set and 1,525 images in test set.
We use camera parameters and disparity maps to calculate depth maps.
The quality of depth data in this dataset is not as good as NYUDv2, and the scenes have wider ranges and more complicated structures.

\section{More experiments of the initialization of introduced parameters}
In Table~\ref{table:ablation_initialization}, we compare more different initialization settings of the introduced parameters $a_k$ and $t$.
When we change the initialization settings, the performance may slightly drop, but still outperforms the baseline and 2.5D convolution.
This validates the effectiveness and the robustness of the learnable depth receptive field in our method.
Small initialization values of $a_k$ seems to have relatively obvious harm on the performance.
We suppose that it is because in this case the receptive fields of different kernels are largely overlapped at the initial state, and it brings difficulty for learning.

\begin{table}[htbp]
  \begin{center}
  \caption{
  Results of different initialization of $a_k$ and $t$.
  The backbone model is ResNet-50
  }
  \label{table:ablation_initialization}
  \setlength{\tabcolsep}{9pt}
  \begin{tabular}{lllccc}
    \hline\noalign{\smallskip}
    Method & $a_k$ & $t$ & mIoU(\%) & pixel Acc(\%)\\
    \noalign{\smallskip}
    \hline
    \noalign{\smallskip}
    Baseline               &  - & - & 44.56  & 73.01 \\
    2.5D\cite{2_5D}        &  - & - & 48.23 & 75.73 \\
    \noalign{\smallskip}
    \hline
    \noalign{\smallskip}
    \multirow{3}{*}{Malleable 2.5D}
     & $[-4,-2,0,2,4]$ &  1 & 48.66 & 75.94 \\
     & $[-2,-1,0,1,2]$ &  1 & \textbf{48.80} & \textbf{76.03} \\
     & $[-1,-0.5,0,0.5,1]$ &  1 & 48.42 & 75.78 \\
    \noalign{\smallskip}
    \hline
    \noalign{\smallskip}
    \multirow{3}{*}{Malleable 2.5D}
     & $[-2,-1,0,1,2]$ &  0.5 & 48.69 & 75.81 \\
     & $[-2,-1,0,1,2]$ &  1 & \textbf{48.80} & \textbf{76.03} \\
     & $[-2,-1,0,1,2]$ &  2 & 48.74 & 75.83 \\
    \hline
  \end{tabular}
  \end{center}
\end{table}

\section{Images of Assigning Functions $h_k$ and $g_k$}
\begin{figure}[htbp]
  \centering
  \subfigure[Initialization of $h_k$]{
  \includegraphics[width=0.3\textwidth]{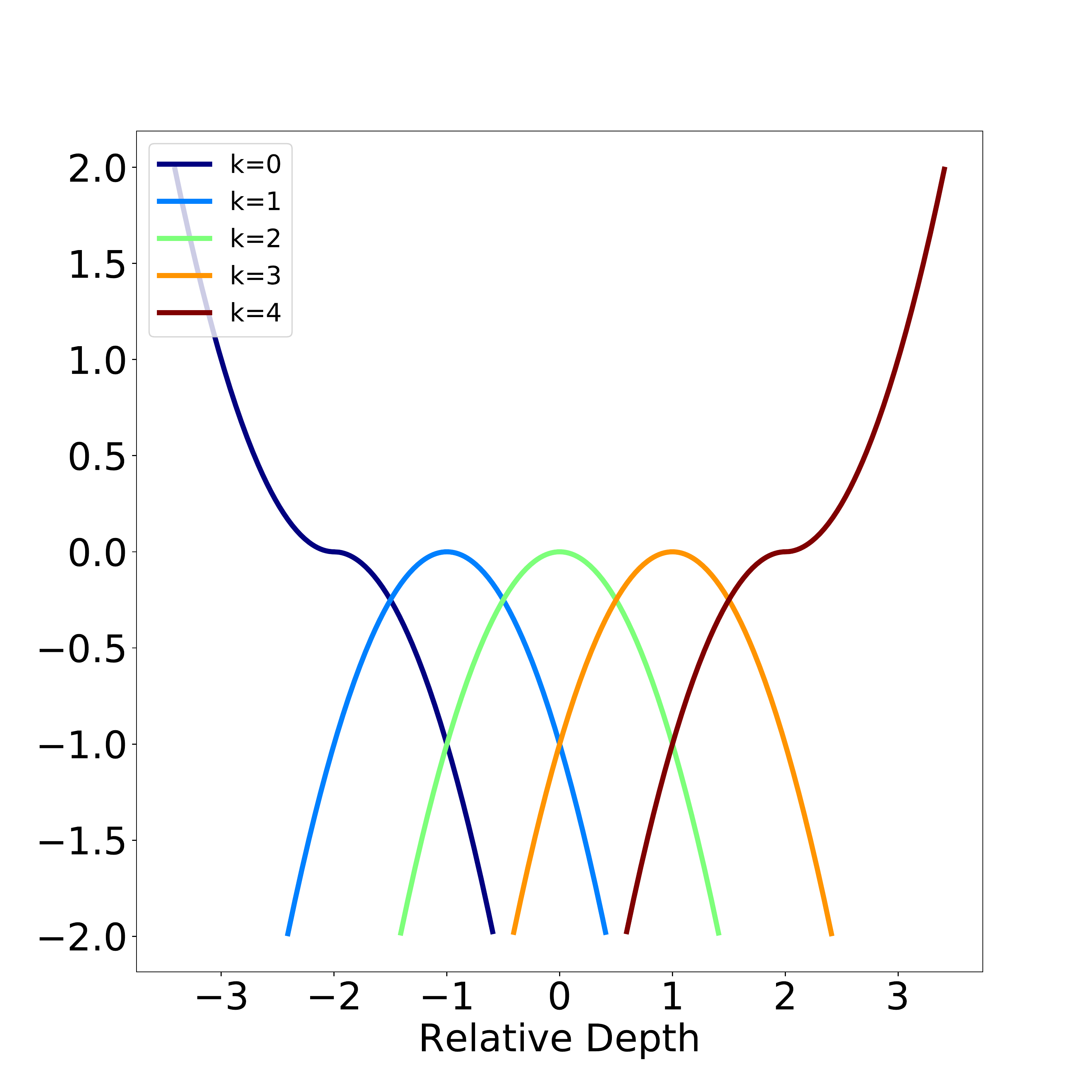}
  }
  \subfigure[Initialization of $g_k$]{
  \includegraphics[width=0.3\textwidth]{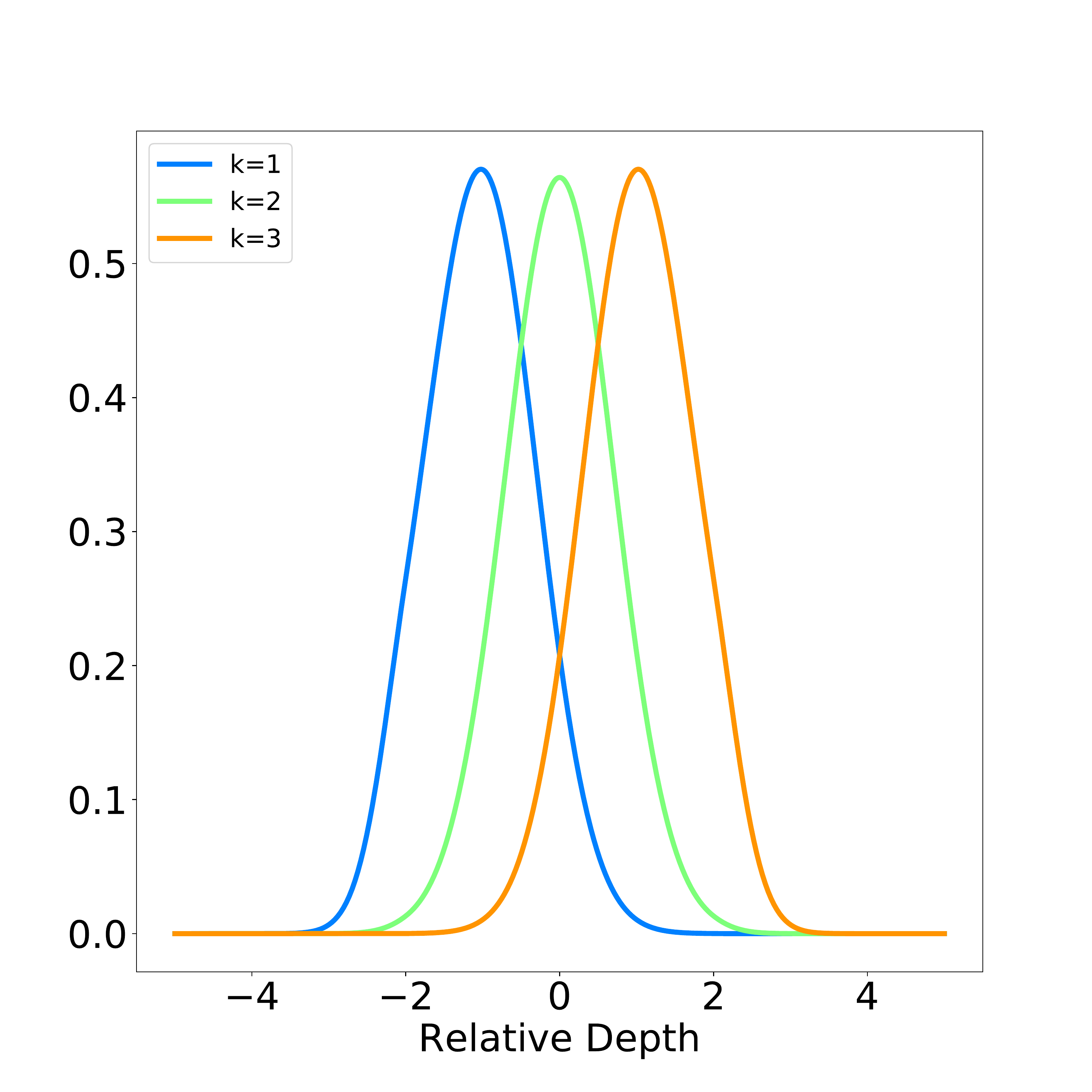}
  }\\
  \subfigure[$h_k$ at res2]{
  \includegraphics[width=0.2\textwidth]{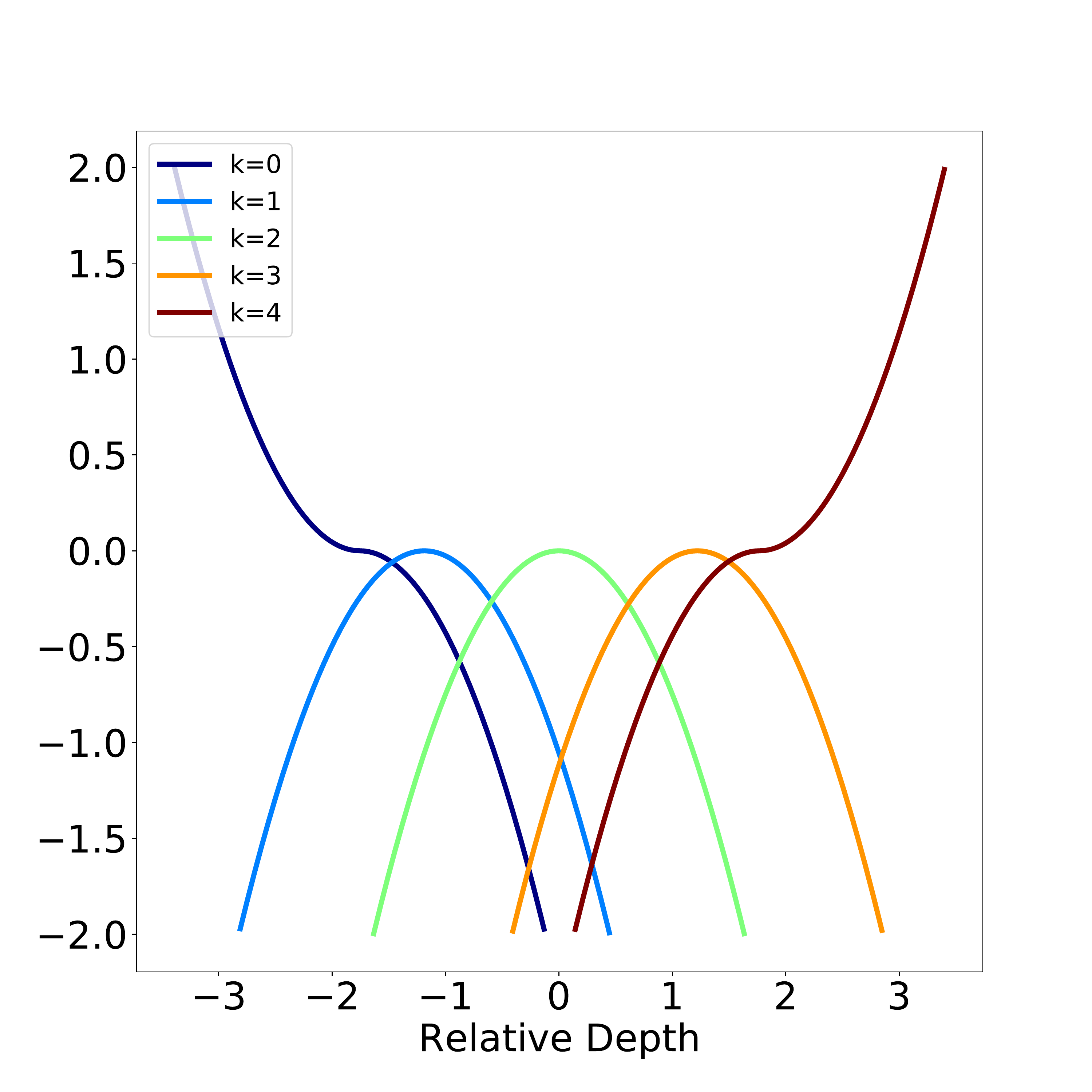}
  }
  \subfigure[$g_k$ at res2]{
  \includegraphics[width=0.2\textwidth]{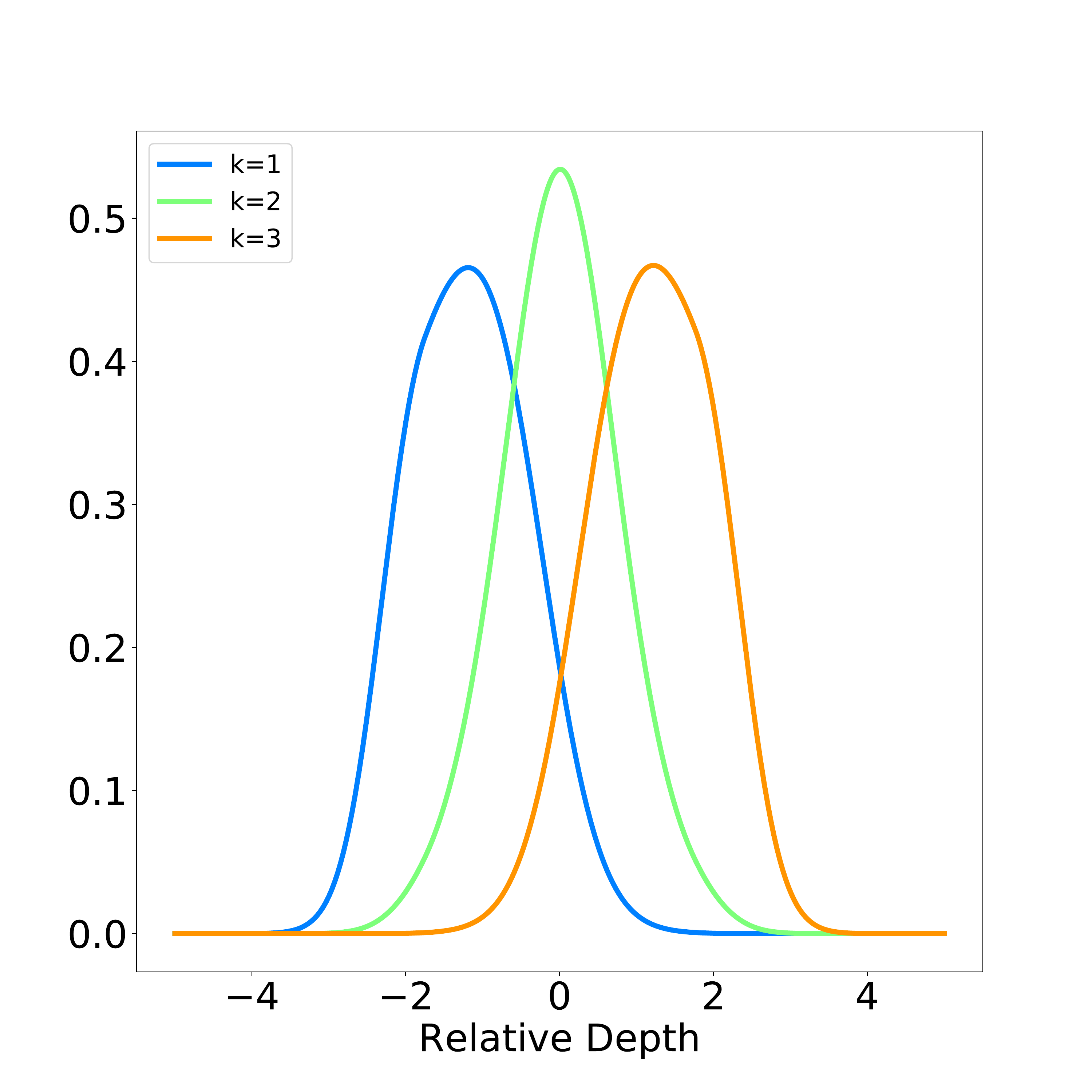}
  }
  \subfigure[$h_k$ at res3]{
  \includegraphics[width=0.2\textwidth]{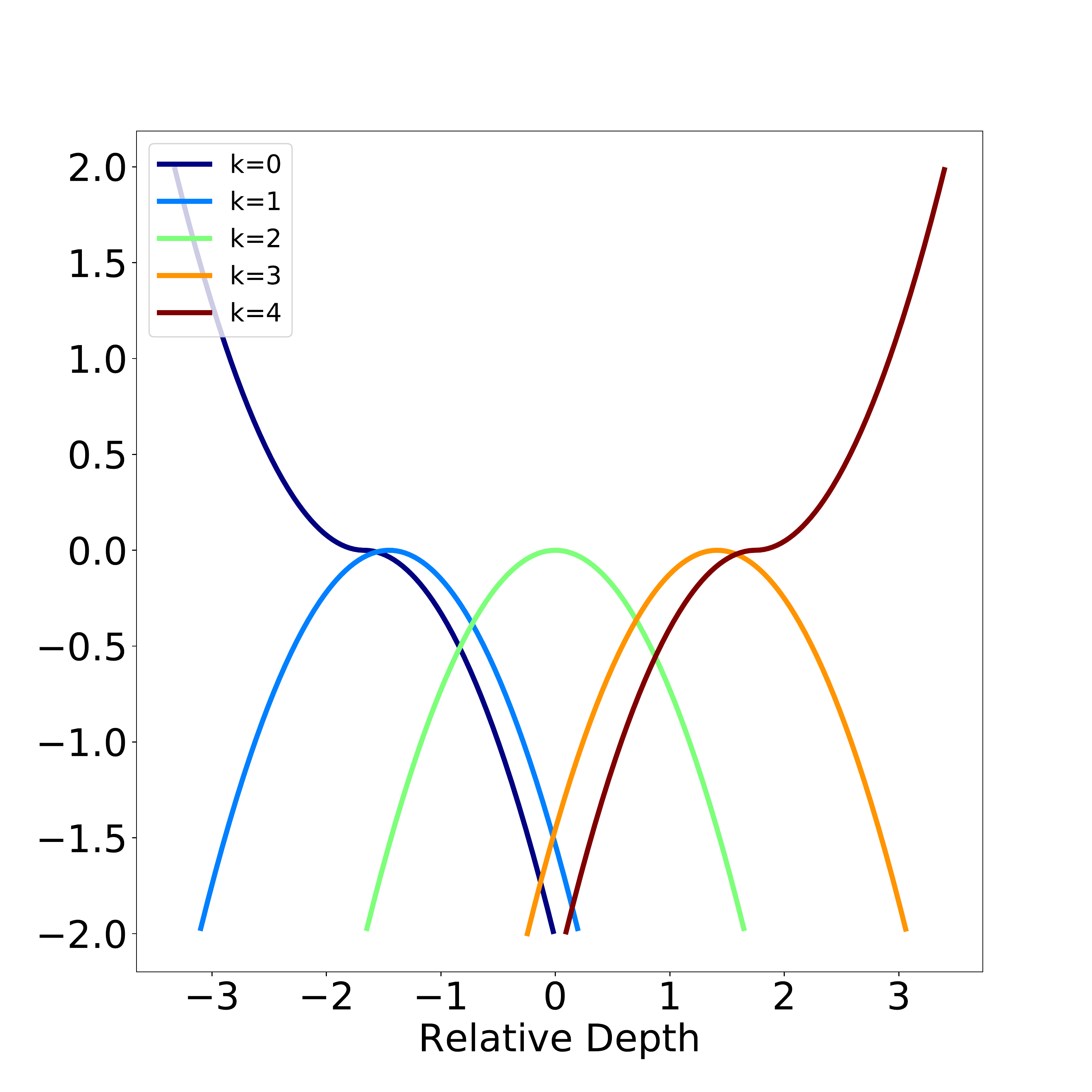}
  }
  \subfigure[$g_k$ at res3]{
  \includegraphics[width=0.2\textwidth]{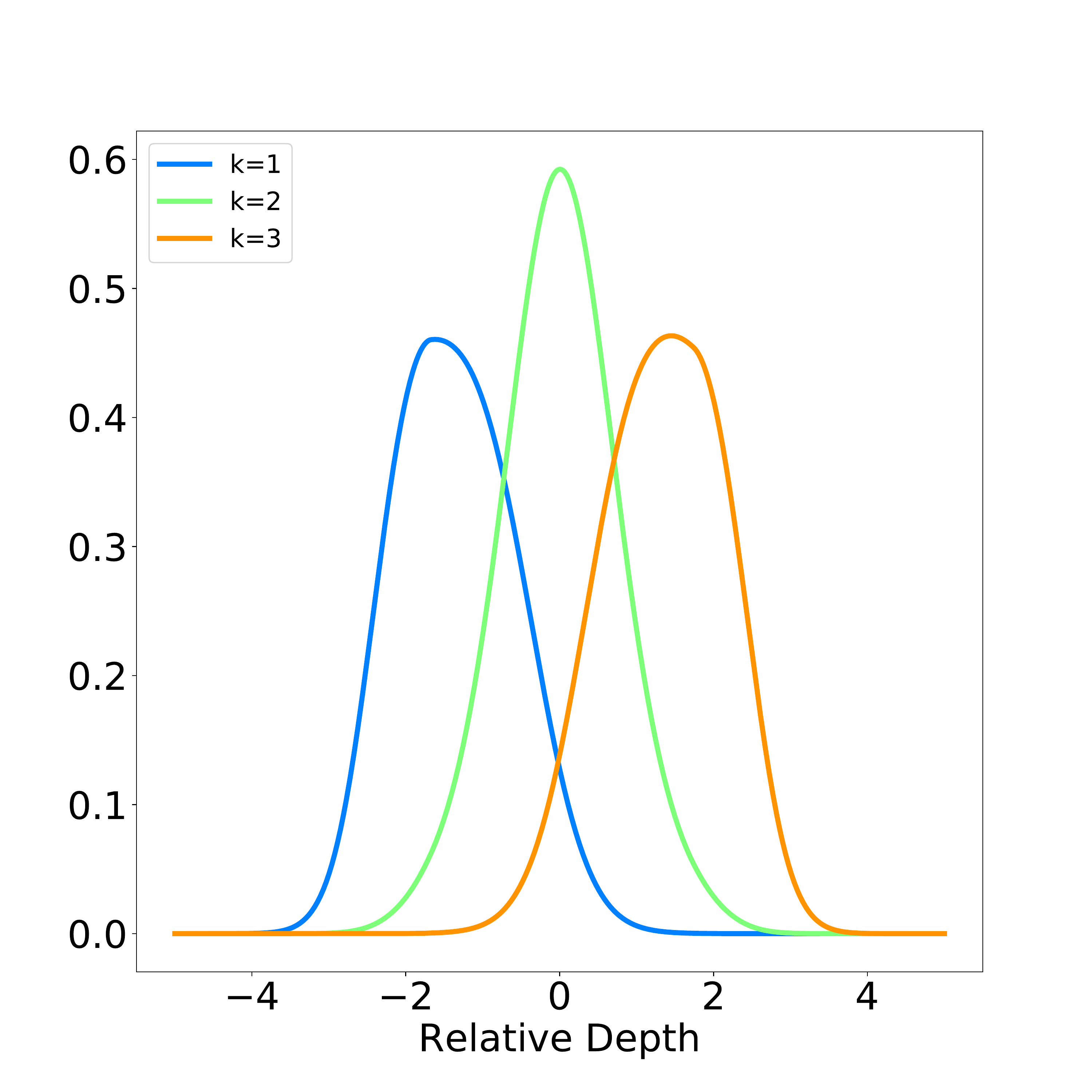}
  }
  \subfigure[$h_k$ at res4]{
  \includegraphics[width=0.2\textwidth]{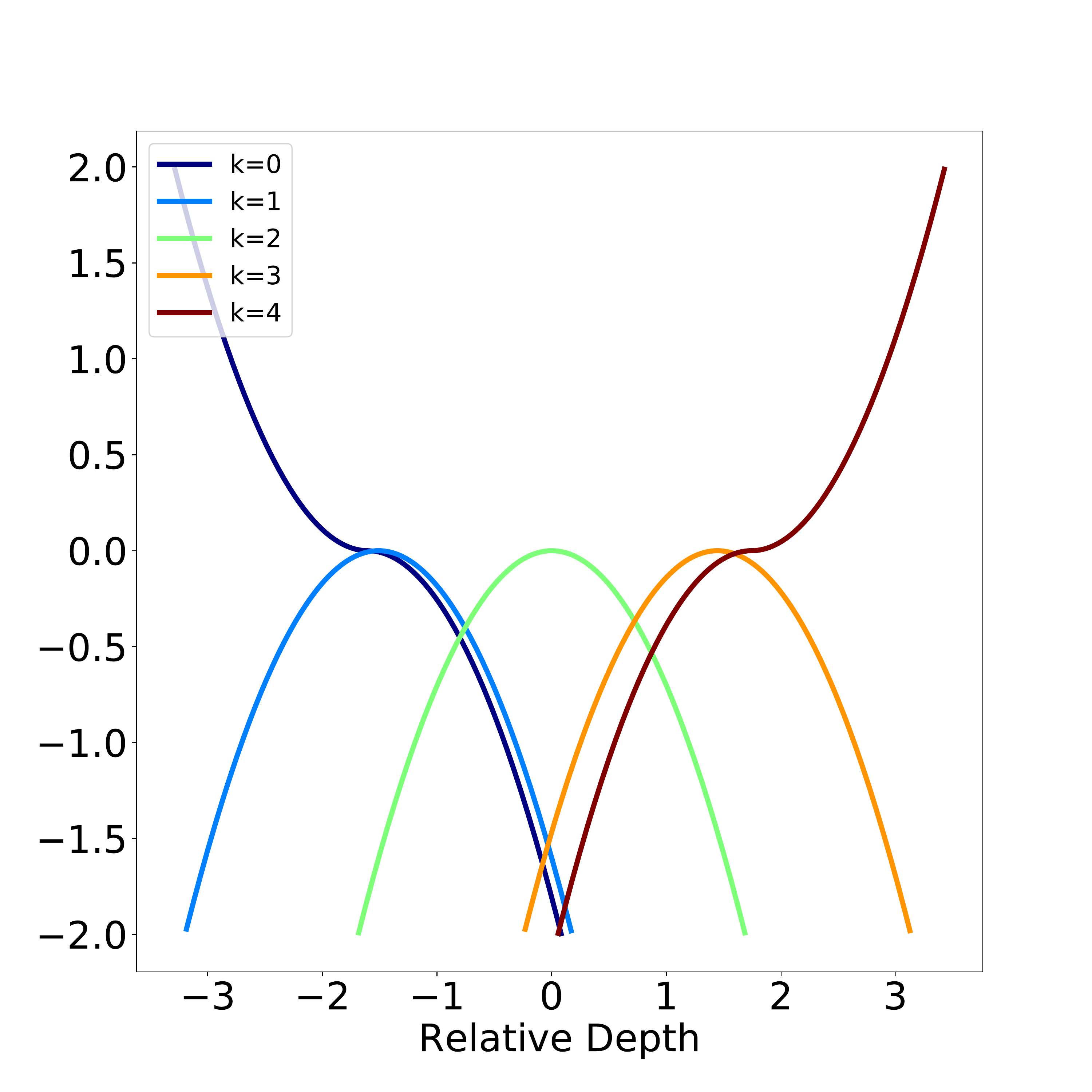}
  }
  \subfigure[$g_k$ at res4]{
  \includegraphics[width=0.2\textwidth]{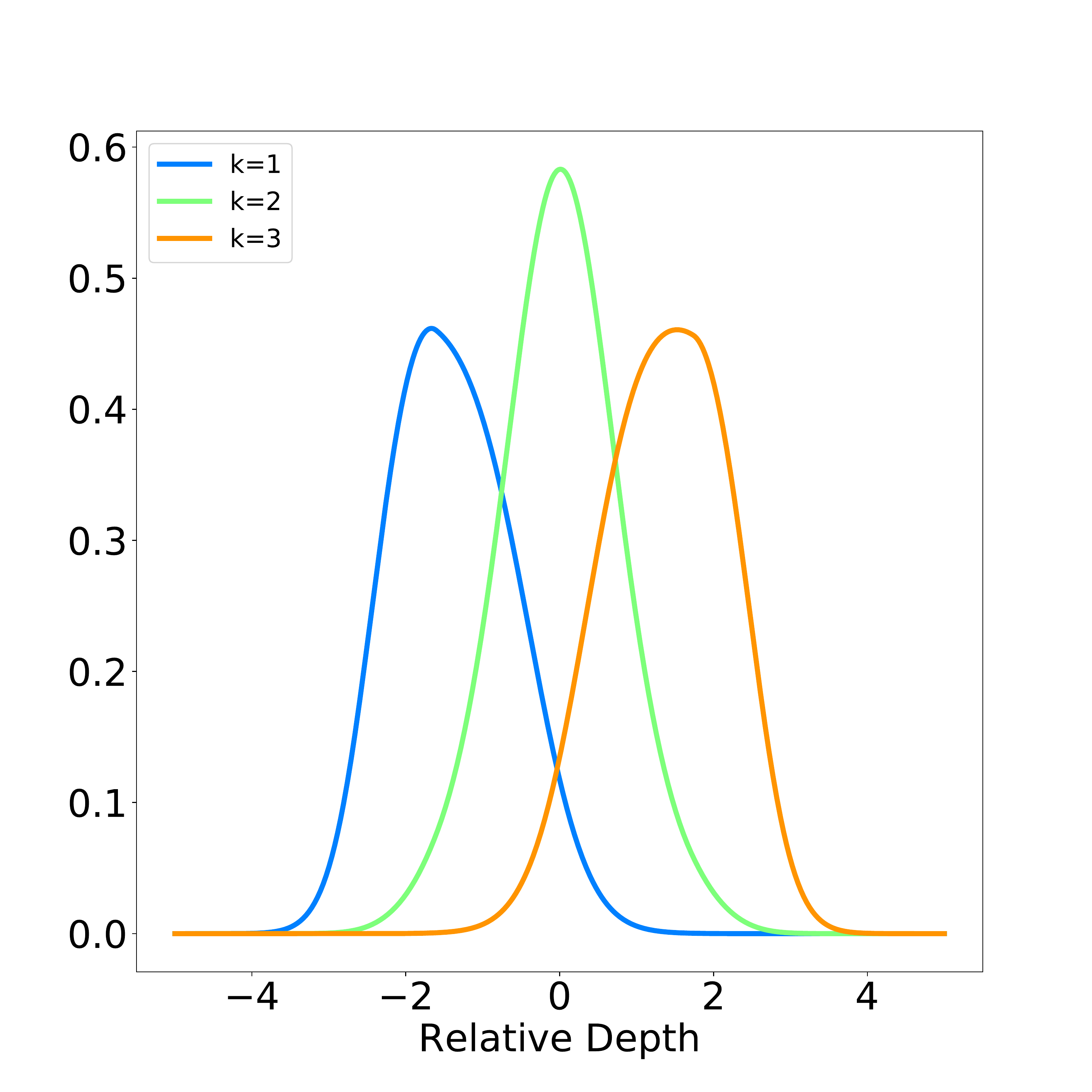}
  }
  \subfigure[$h_k$ at res5]{
  \includegraphics[width=0.2\textwidth]{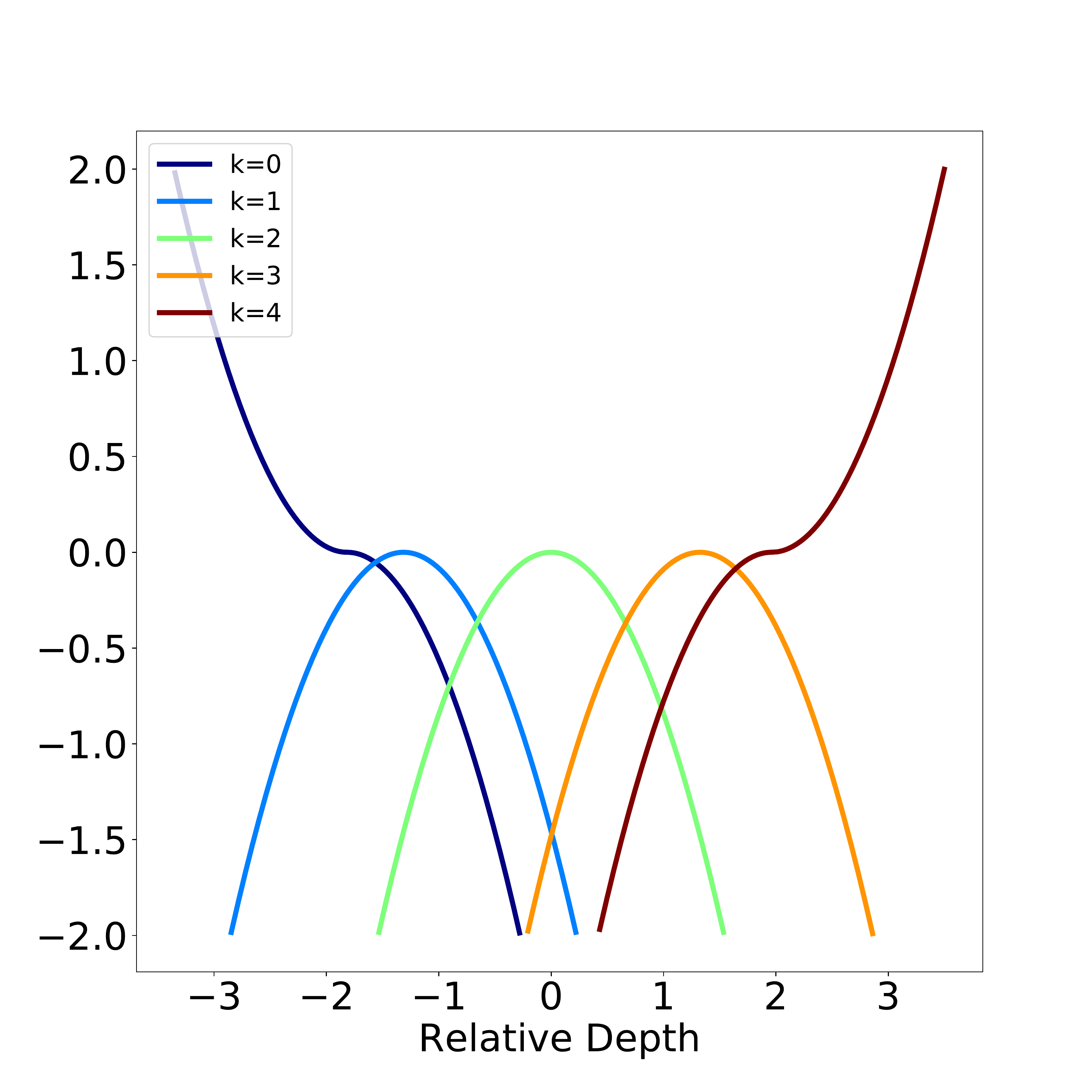}
  }
  \subfigure[$g_k$ at res5]{
  \includegraphics[width=0.2\textwidth]{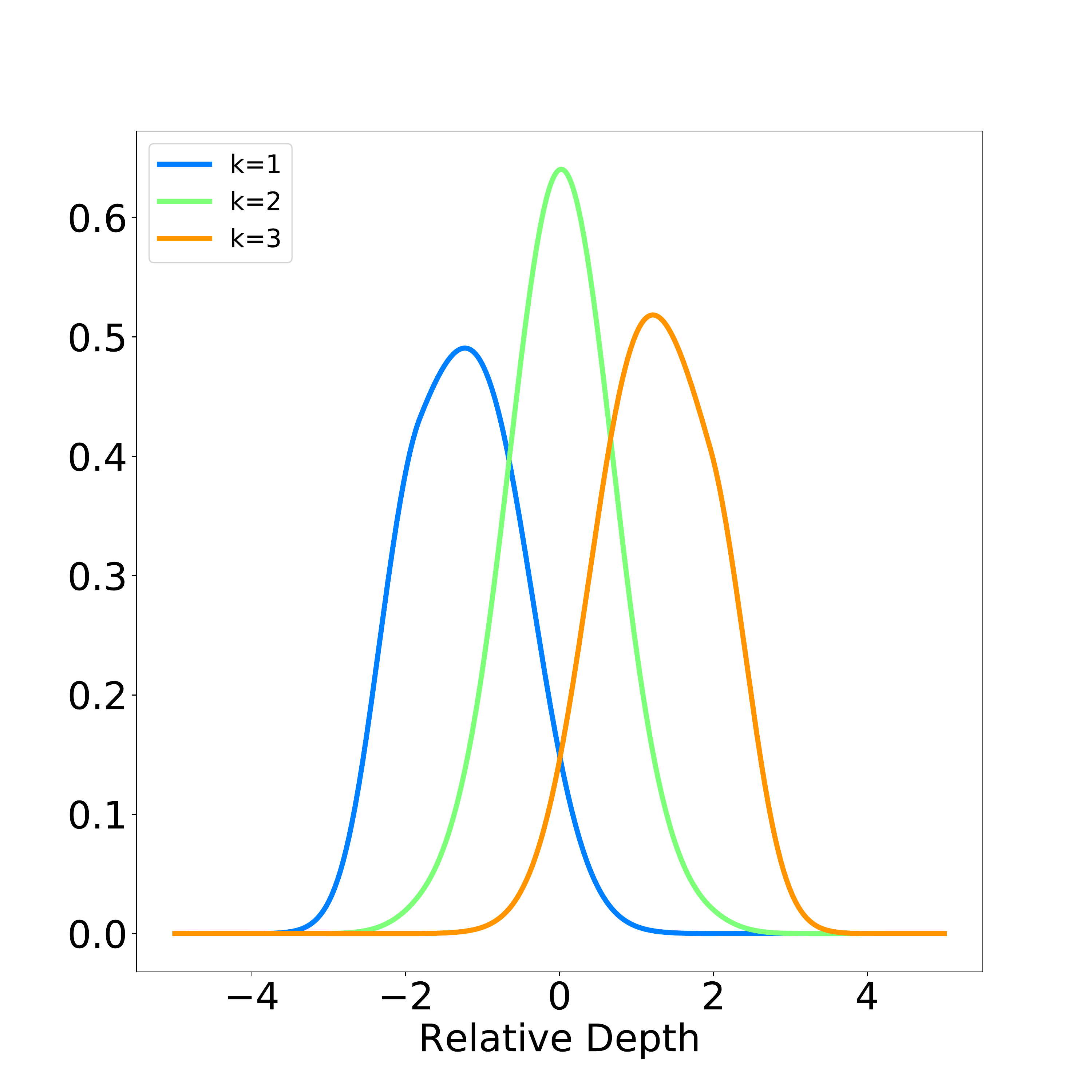}
  }
  \caption{
    Images of $h_k$ and $g_k$.
    (a) and (b) are the initialization of $h_k$ and $g_k$.
    And the rest subfigures are $h_k$-s and $g_k$-s of each malleable 2.5D convolution at different ResNet stages after training
  }
  \label{fig:h_g}
\end{figure}

To give a clear illustration of the assigning functions $h_k$ and $g_k$, we present several real-case images of $h_k$ and $g_k$ in Fig.~\ref{fig:h_g}.
We draw the images of the initialization state of $h_k$ and $g_k$, and we also draw the images of $h_k$ and $g_k$ in a model trained on NYUDv2.

\section{Comparisons of Depth Receptive Functions}
\begin{figure}[htbp]
  \centering
  \includegraphics[width=0.85\textwidth]{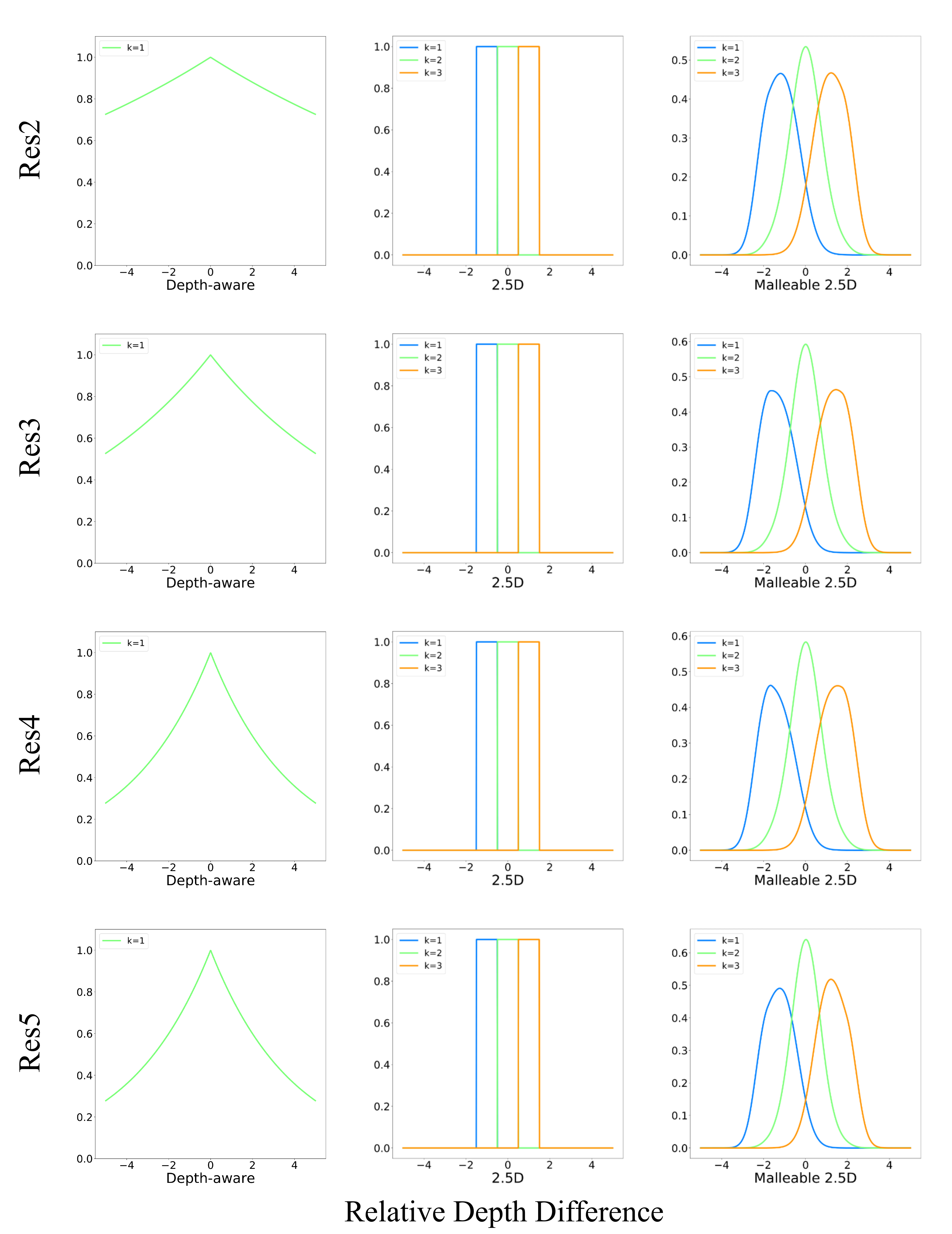}
  \caption{
  Comparison of depth receptive field functions $g_k$ on NYUDv2 where the depth $\mathbf{d}(\mathbf{c}_i)=1m$.
  We compare depth-aware, 2.5D and malleable 2.5D convolution at each stages of ResNet after training.
  Note that we scale the y-axis to see better.
  The overall scale does not affect output results because of batch normalizations
  }
  \label{fig:rf_compare_nyu}
\end{figure}

\begin{figure}[htbp]
  \centering
  \includegraphics[width=0.85\textwidth]{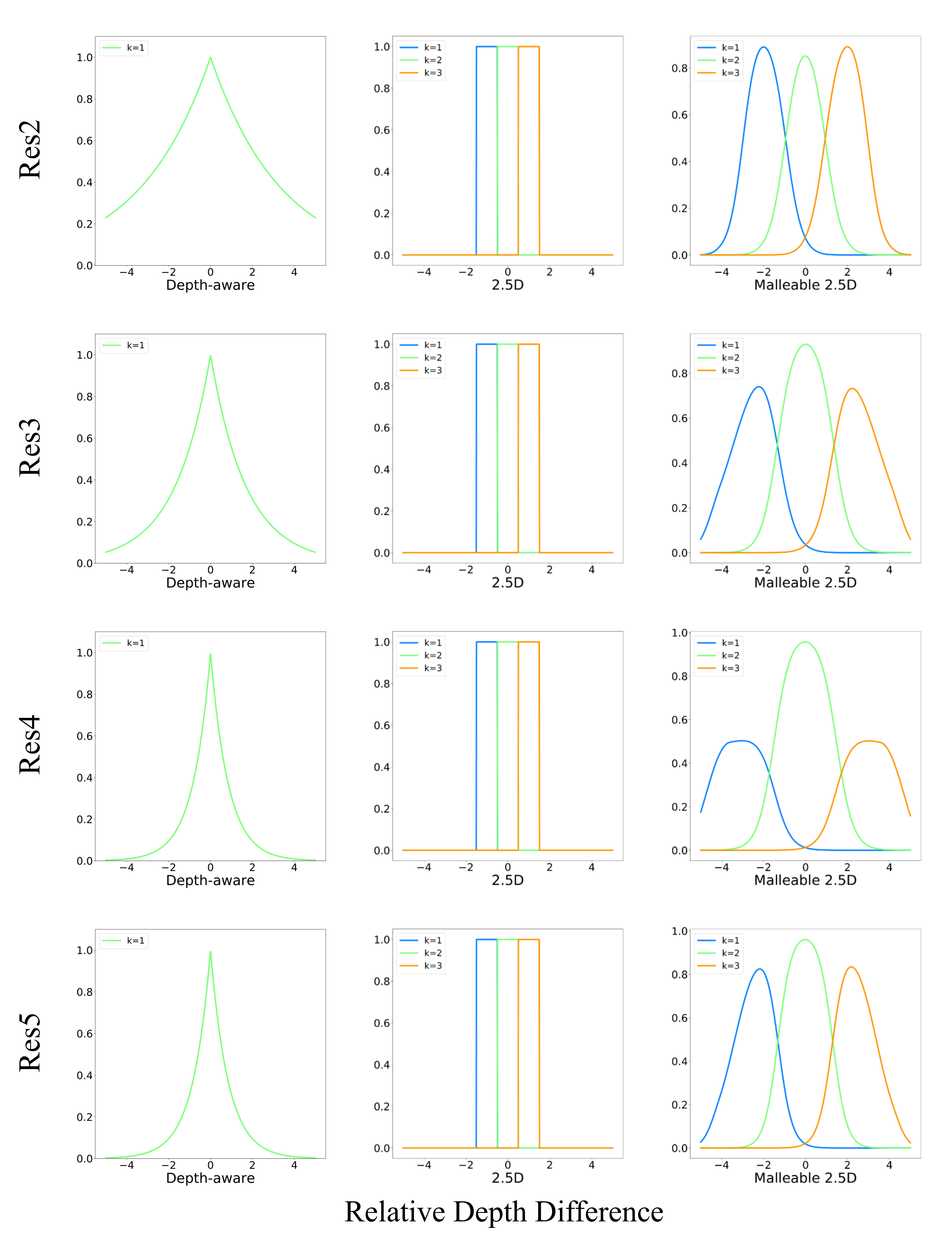}
  \caption{
  Comparison of depth receptive field functions $g_k$ on Cityscapes where the depth $\mathbf{d}(\mathbf{c}_i)=20m$.
  We compare depth-aware, 2.5D and malleable 2.5D convolution at each stages of ResNet after training.
  Note that we scale the y-axis to see better.
  The overall scale does not affect output results because of batch normalizations
  }
  \label{fig:rf_compare_city}
\end{figure}

In Fig.~\ref{fig:rf_compare_nyu} and Fig.~\ref{fig:rf_compare_city}, we compare depth receptive fields of depth-aware, 2.5D and malleable 2.5D convolution in NYUDv2 and Cityscapes.
In NYUDv2, pixels are closer in 3D space than those in Cityscapes since they are respectively indoor and outdoor scenes.
Therefore, it is intuitive that convolutions should have larger depth receptive fields on Cityscapes than NYUDv2.
From the figures we can see that malleable 2.5D convolution indeed learns wider depth receptive fields for Cityscapes, while depth-aware and 2.5D convolutions cannot automatically fit different environments.

\section{Effects of Kernel Rebalancing}
In Fig.~\ref{fig:rebalance}, we present the effects of kernel rebalancing.
We show the kernel rebalancing results of all four malleable 2.5D convolutions in the model trained on NYUDv2.
As we know,
in earlier stages, local geometric features play a more important role.
and in later stages, the importance of capturing context increases.
The rebalancing parameters in early stages only fix part of the imbalance problem and keep the two further kernels decayed compared to the center kernel, which makes the convolution sensitive to local depth changes.
When comes to the later stages, the rebalancing parameters tend to balance the kernels well and therefore let the convolution able to handle long-distance relations.

\begin{figure}[htbp]
  \centering
  \subfigure[Before rebalance (at res2)]{
  \includegraphics[width=0.33\textwidth]{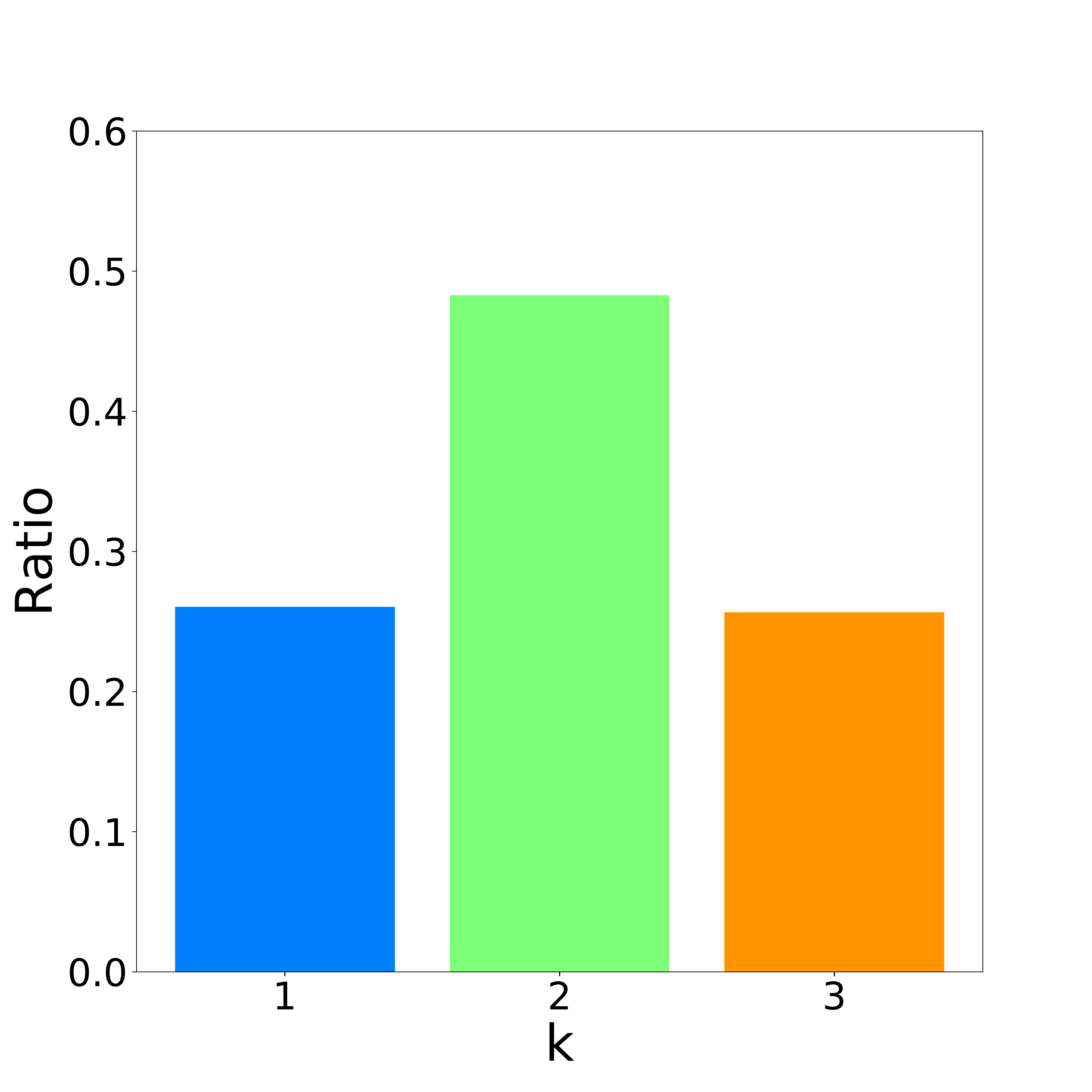}
  }
  \subfigure[After rebalance (at res2)]{
  \includegraphics[width=0.33\textwidth]{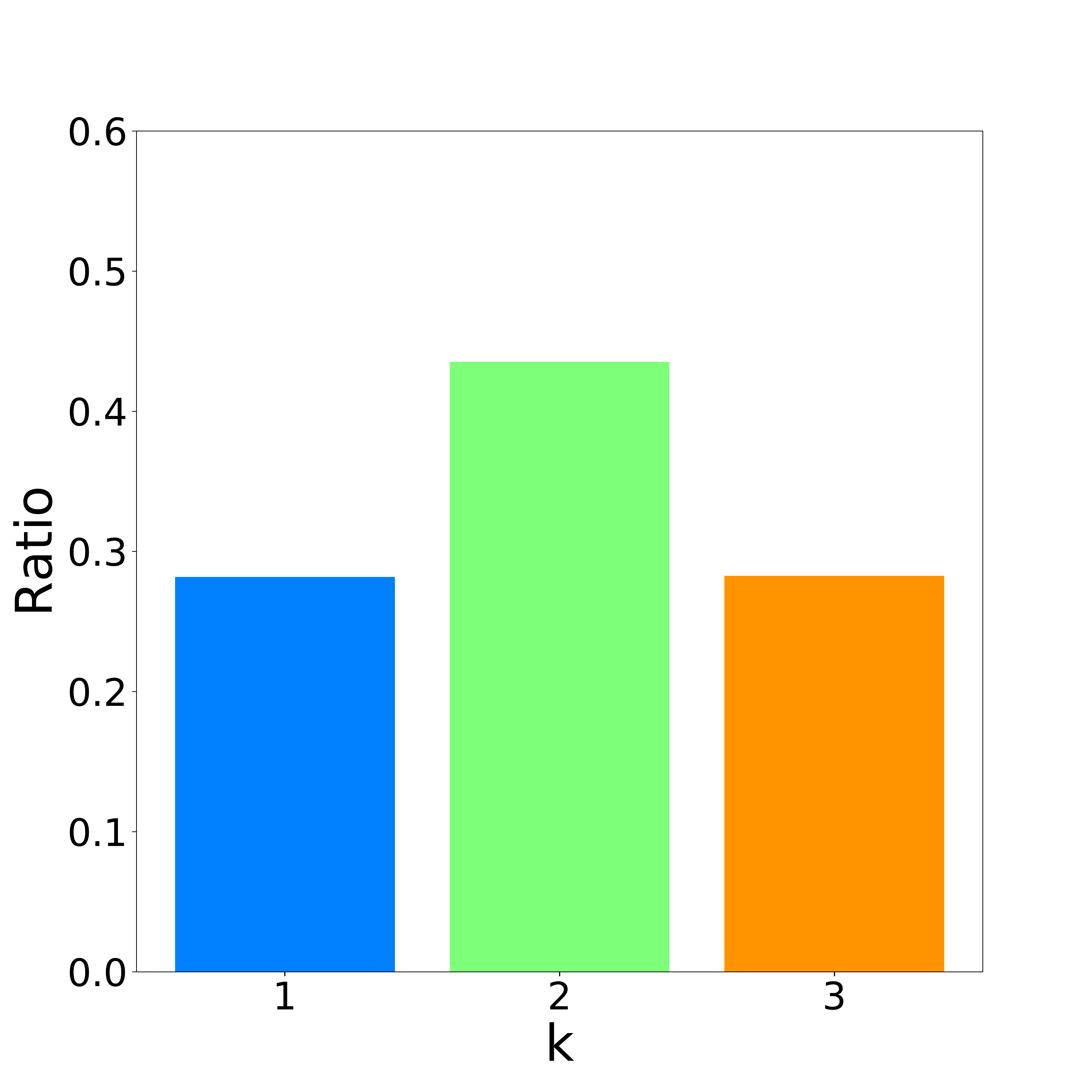}
  }\\
  \subfigure[Before rebalance (at res3)]{
  \includegraphics[width=0.33\textwidth]{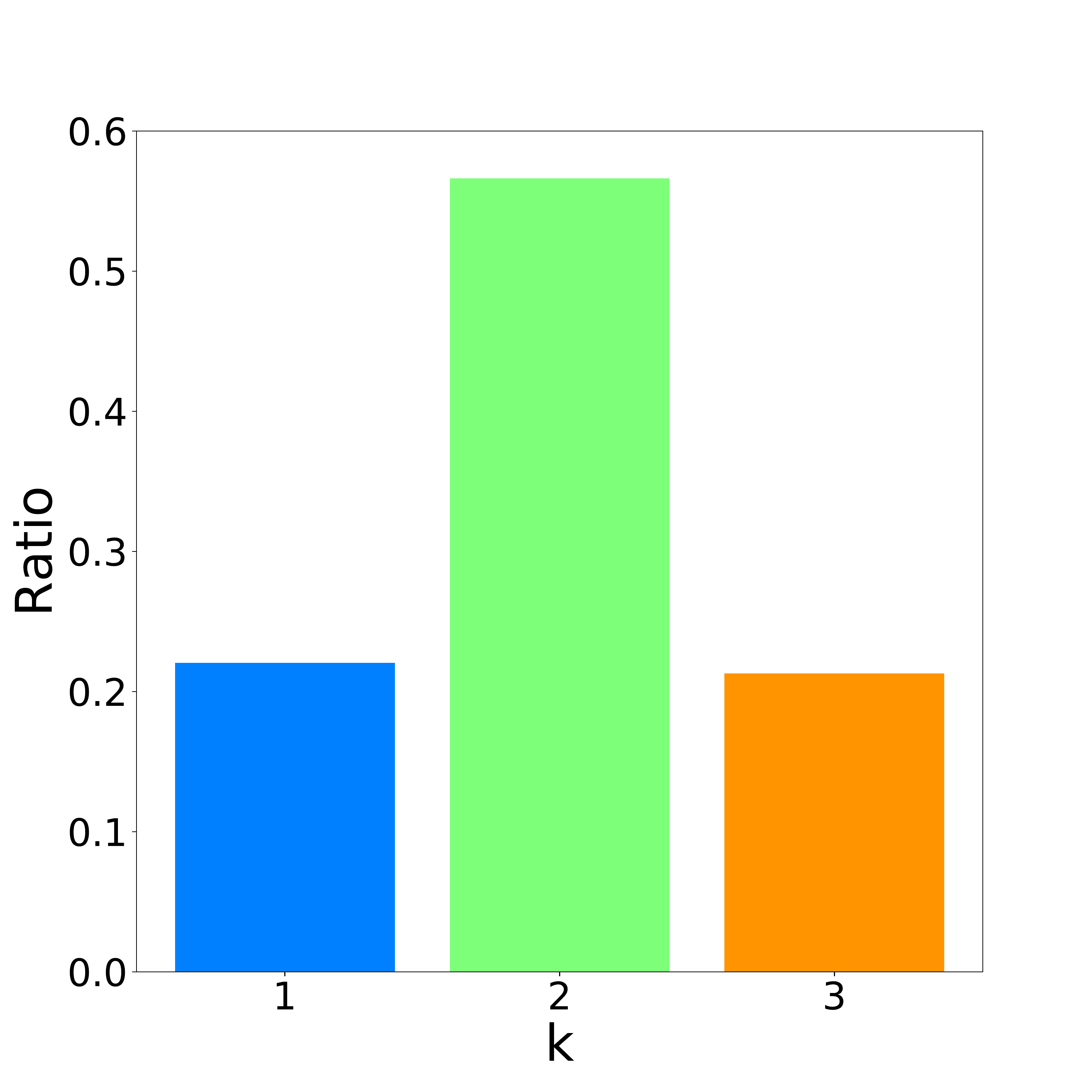}
  }
  \subfigure[After rebalance (at res3)]{
  \includegraphics[width=0.33\textwidth]{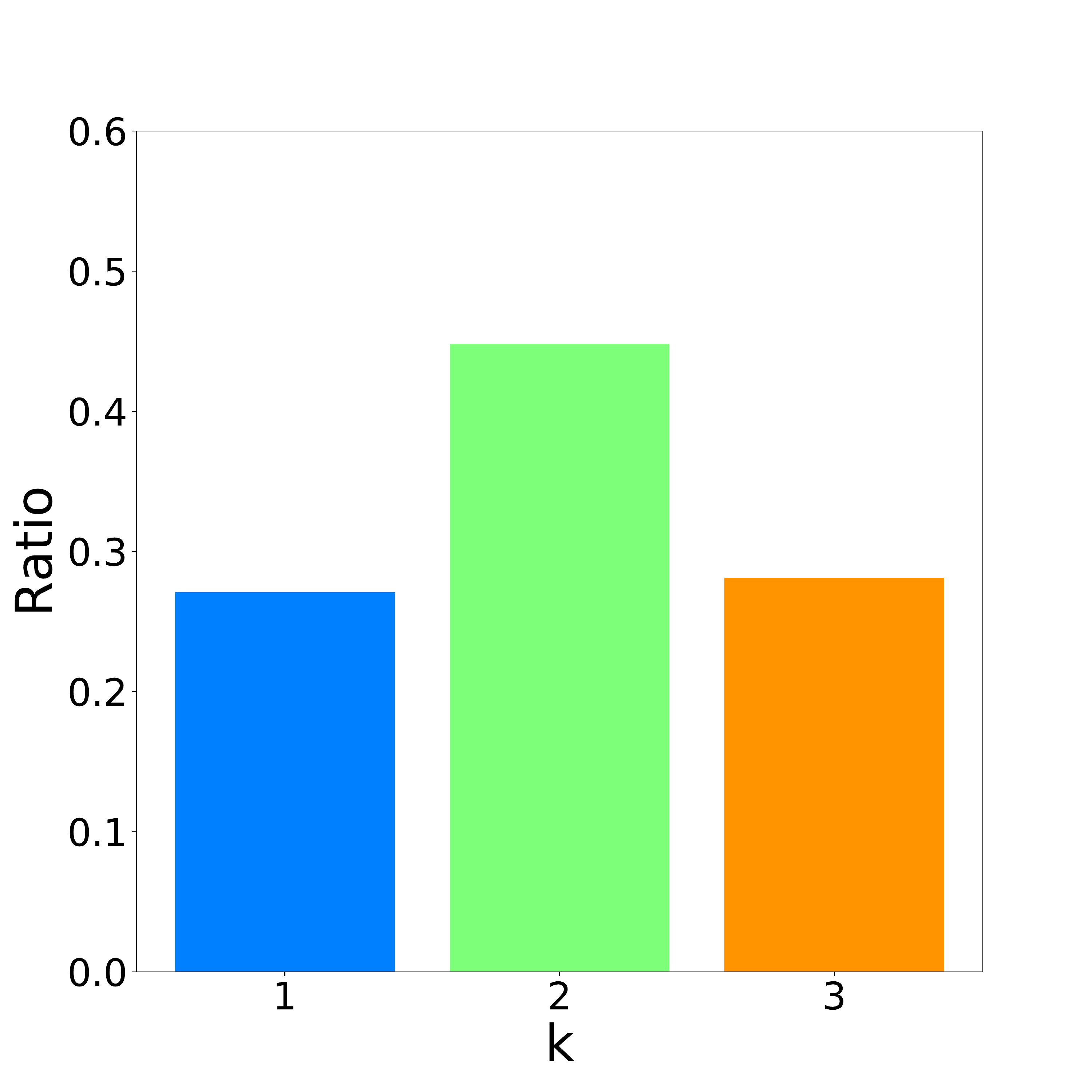}
  }\\
  \subfigure[Before rebalance (at res4)]{
  \includegraphics[width=0.33\textwidth]{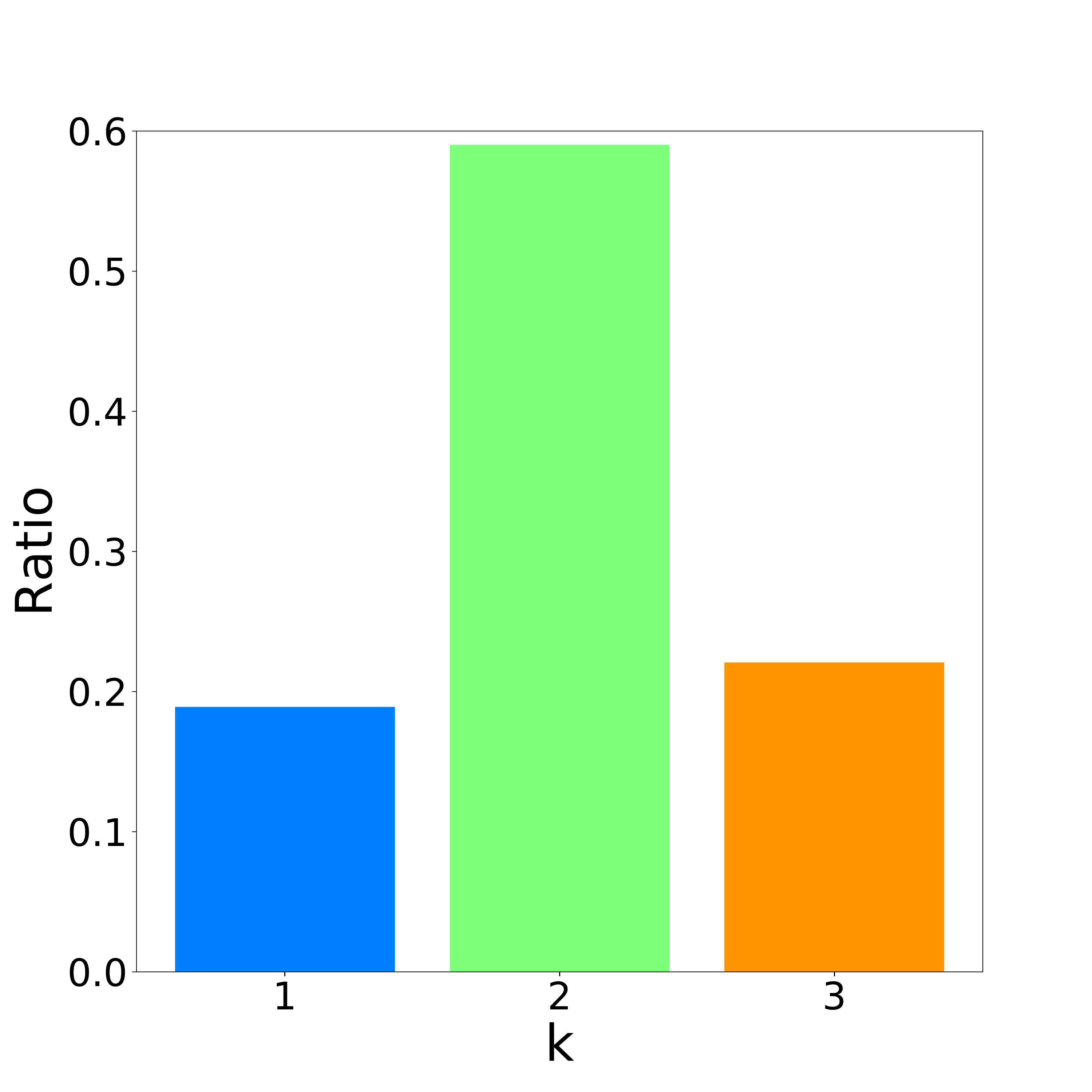}
  }
  \subfigure[After rebalance (at res4)]{
  \includegraphics[width=0.33\textwidth]{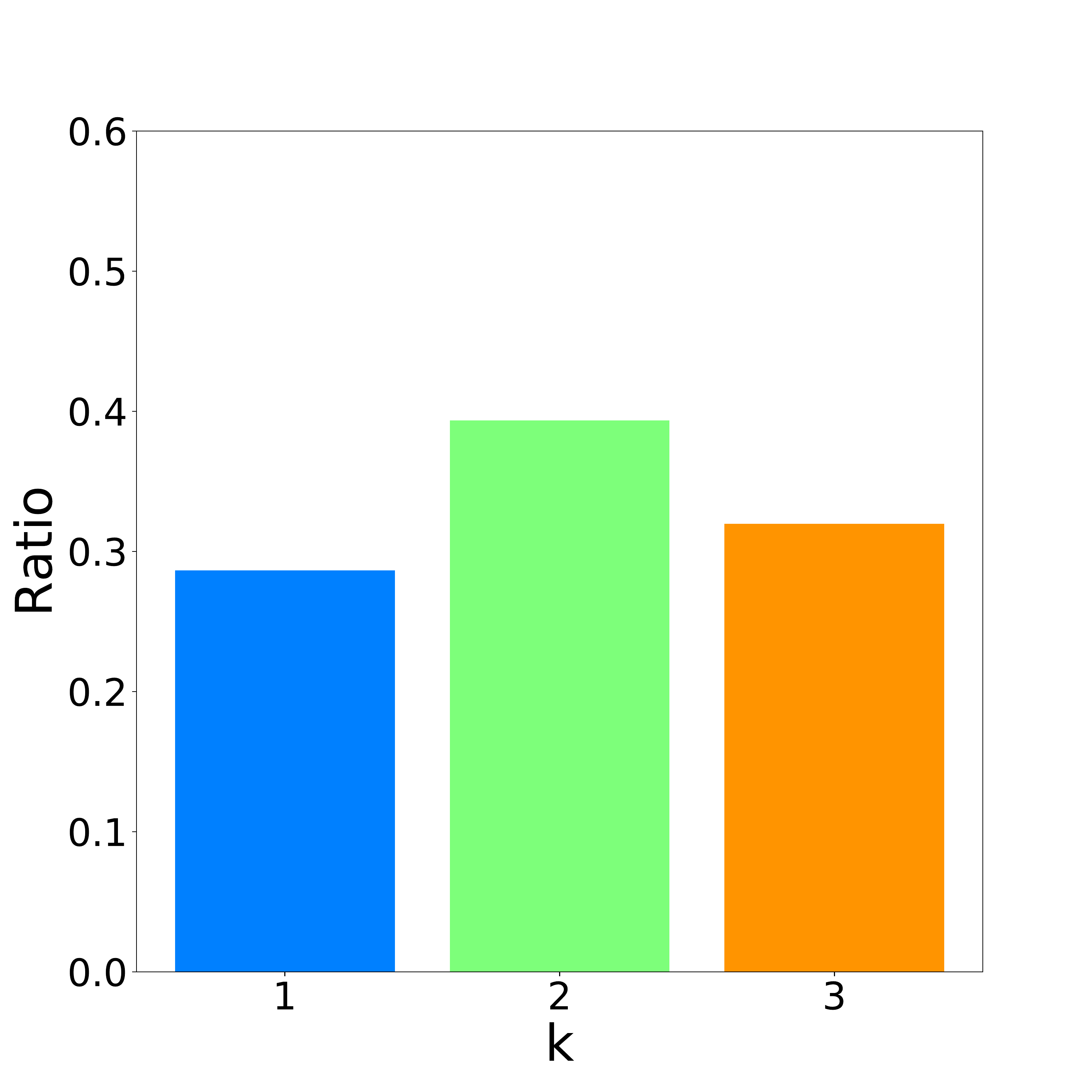}
  }\\
  \subfigure[Before rebalance (at res5)]{
  \includegraphics[width=0.33\textwidth]{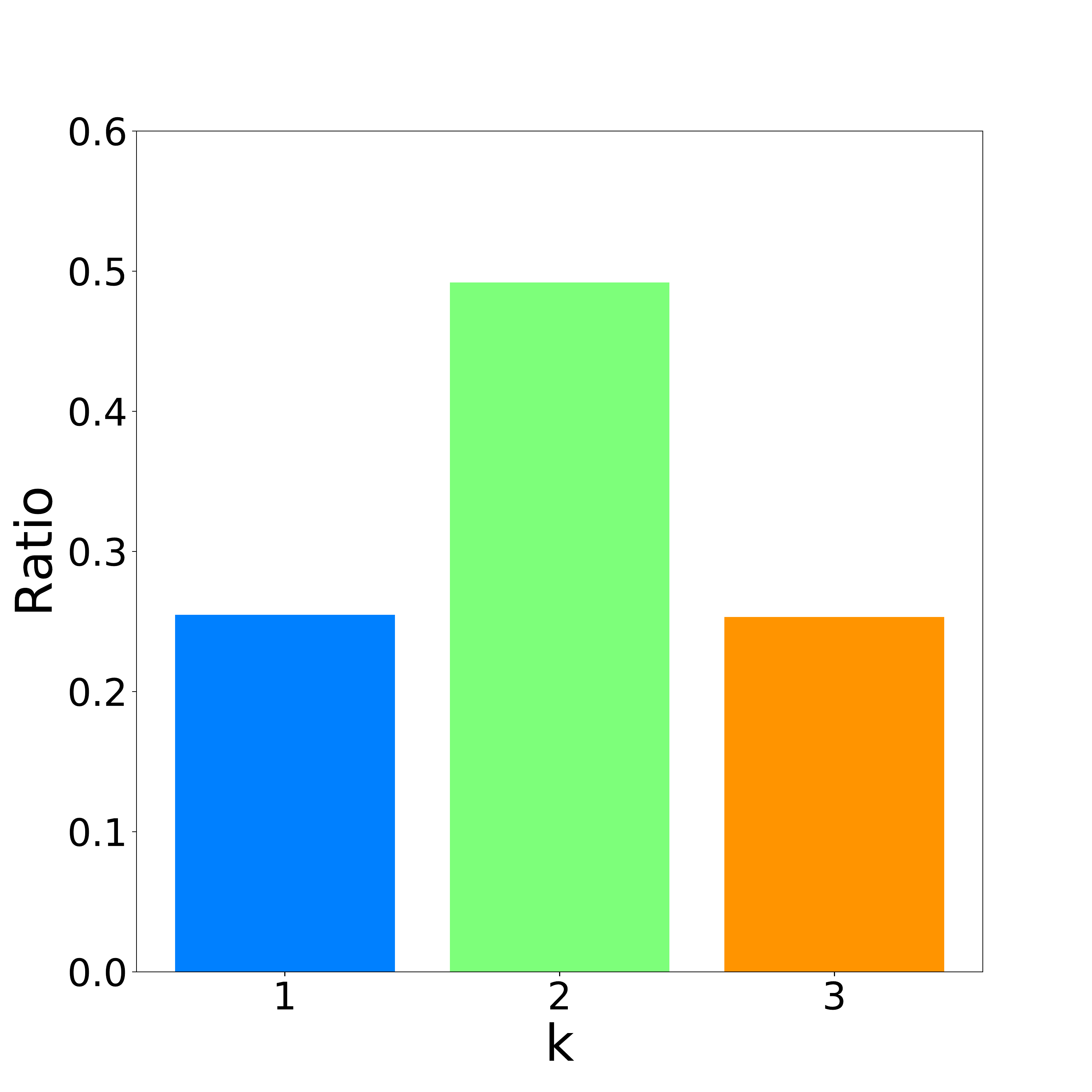}
  }
  \subfigure[After rebalance (at res5)]{
  \includegraphics[width=0.33\textwidth]{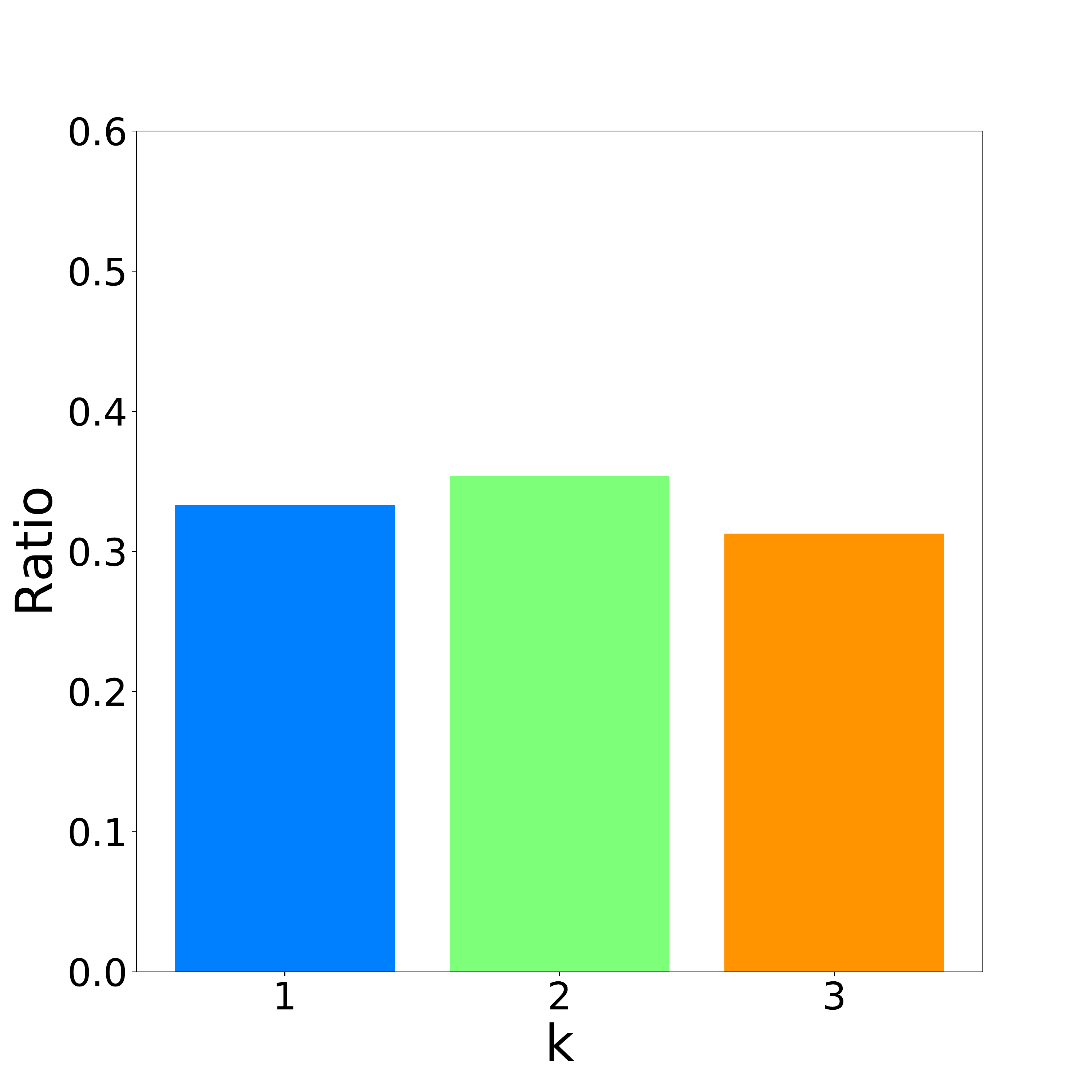}
  }\\
  \caption{
    The ratio of pixels assigned to each kernel, before and after rebalance.
    We count the sum of $g_k$ and $s_k\cdot g_k$ for each kernel across the whole NYUDv2 dataset and calculate the ratio.
  }
  \label{fig:rebalance}
\end{figure}

\section{Visualization of feature maps}
\begin{figure}[htbp]
  \centering
  \includegraphics[width=0.83\textwidth]{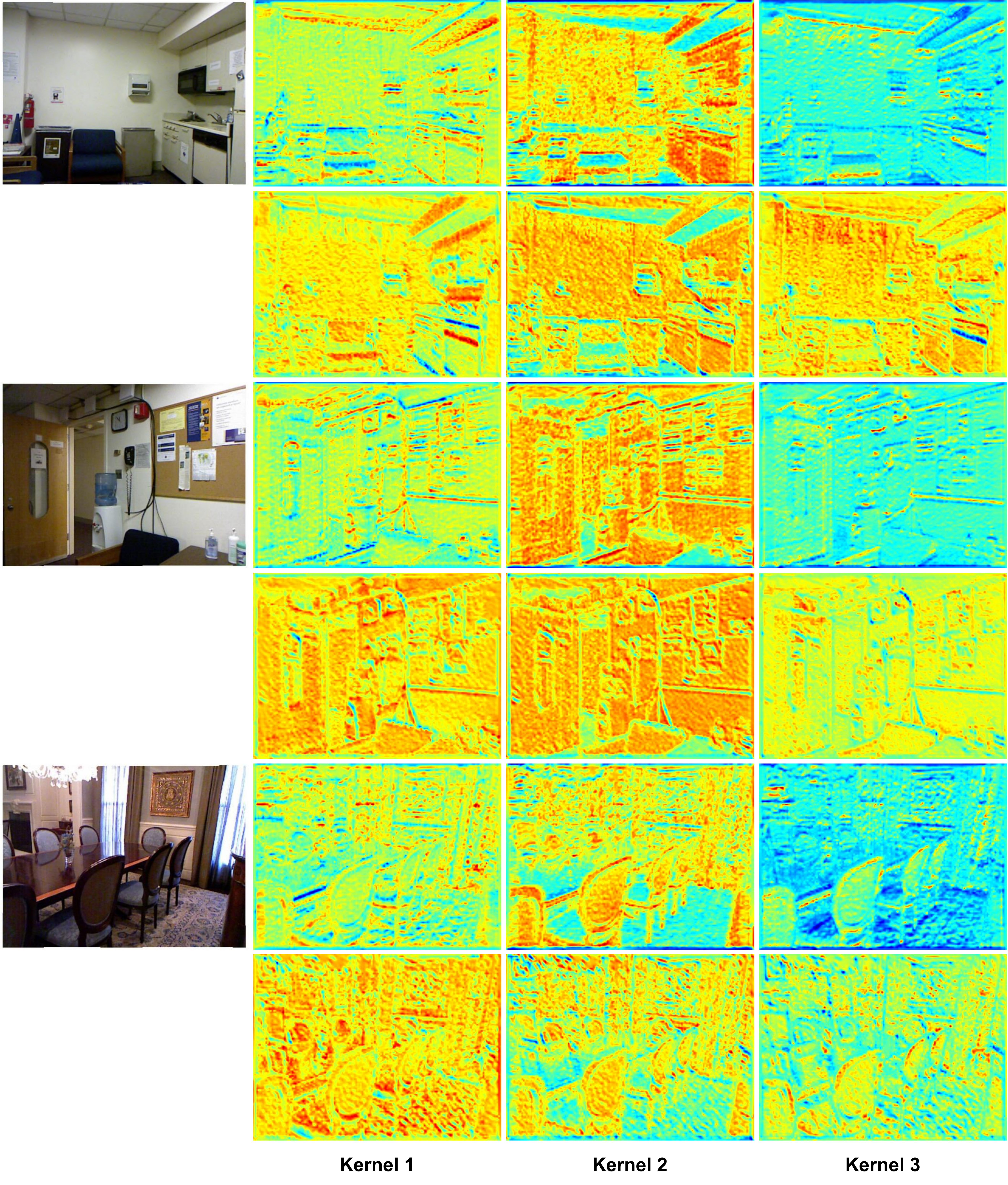}
  \caption{
  Visualization of the feature maps generated by different kernels in a malleable 2.5D convolution.
  We draw 2 feature maps for each kernel and each input image.
  The feature maps of "kernel 1" are determined by what is in front of the center pixel of a local receptive field.
  The feature maps of "kernel 2" are determined by what is around the same depth with the center pixel of a local receptive field.
  The feature maps of "kernel 3" are determined by what is behind the center pixel of a local receptive field.
  }
  \label{fig:features}
\end{figure}

In Fig.~\ref{fig:features}, we visualize the feature maps generated by different kernels of a malleable 2.5D convolution.
We save the output feature maps of the malleable 2.5D convolution in res2 stage of a trained ResNet-101-based model, and select two channels to draw figures.
Generally, the three kernels respectively handle pixels that are "in front of", "around the same depth with" and "behind" the center pixel of a local receptive field.
From the feature maps, we can see that the three kernels indeed learn different relations and can activate accordingly.

\section{Network Structures}
\begin{figure}[htbp]
  \centering
  \subfigure[Network structure for Cityscapes]{
  \includegraphics[width=0.95\textwidth]{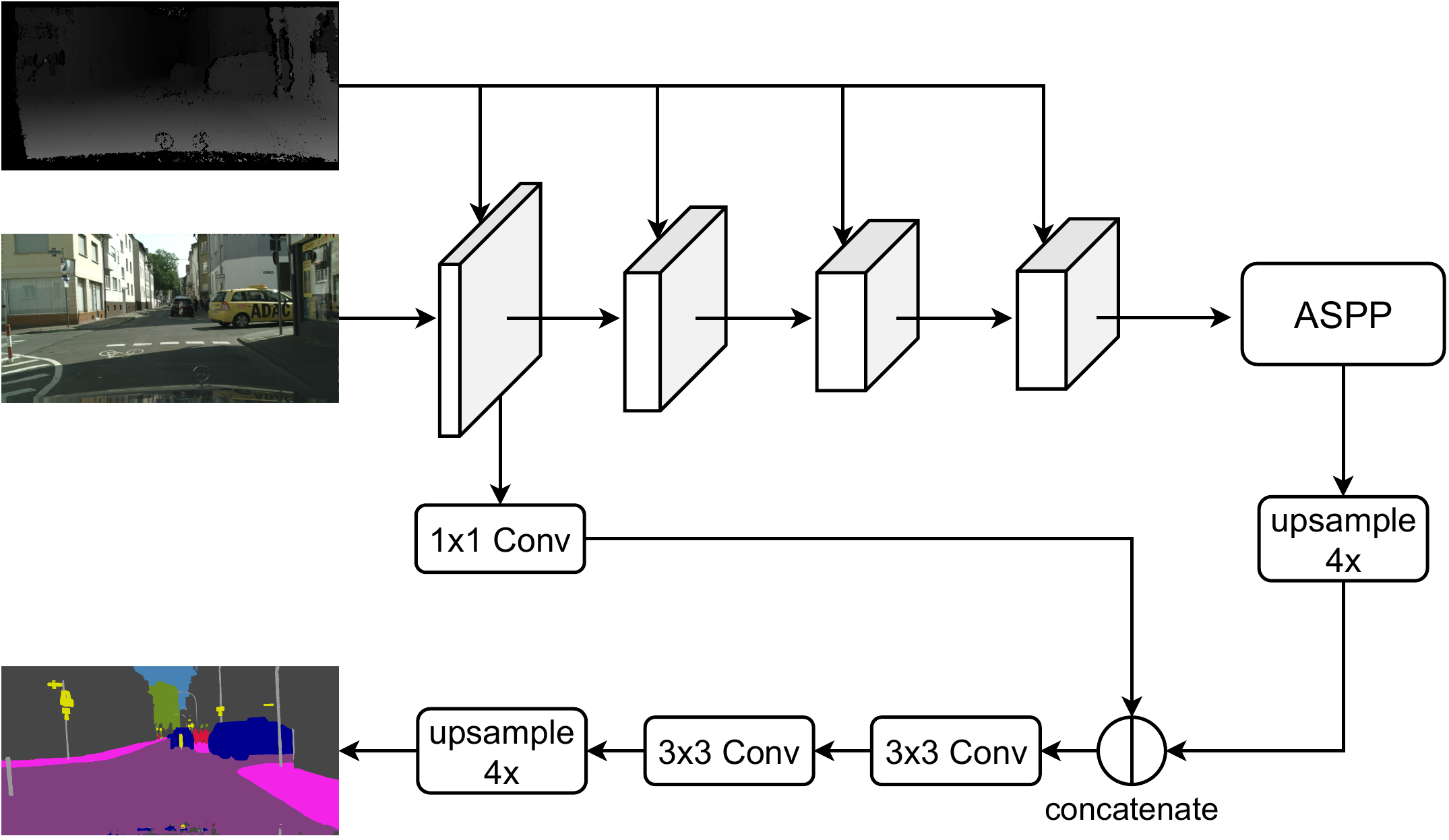}
  }\\
  \subfigure[Network structure for NYUDv2]{
  \includegraphics[width=0.95\textwidth]{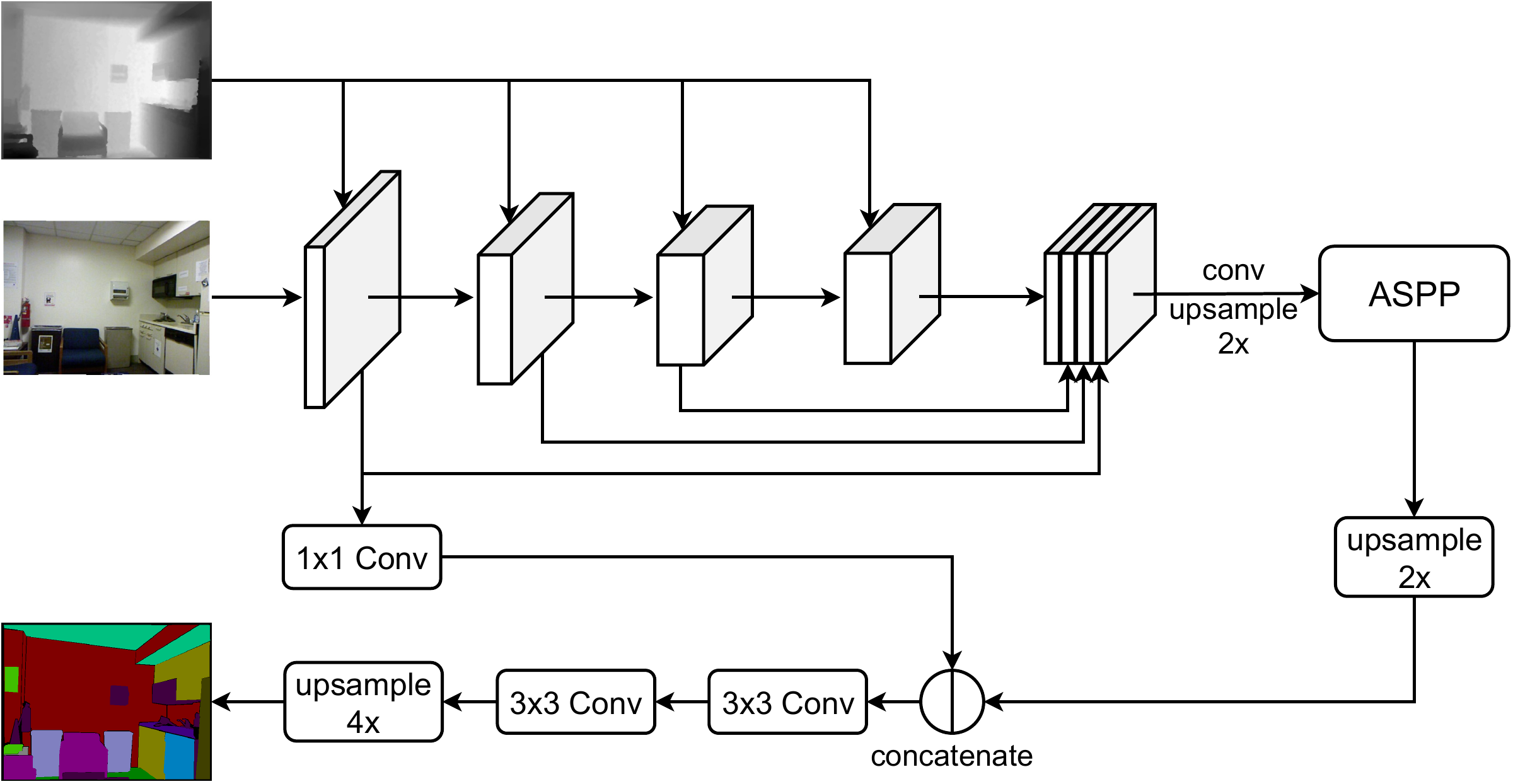}
  }
  \caption{
    Network structures
  }
  \label{fig:network_structure}
\end{figure}

In Fig.~\ref{fig:network_structure}, we present the network structures we use in NYUDv2 and Cityscapes respectively.
We adopt ResNet-based DeepLabv3+ as our baseline network.
To evaluate the effect of our method, we replace the $3\times3$ convolution with a malleable 2.5D convolution in the first residual unit in each stage of the ResNet.
For the NYUDv2 dataset, we adopt a multi-stage merging block on the backbone network.
And for Cityscapes, we keep the original DeepLabv3+ structure.

\end{document}